\documentclass{article}

\usepackage{amsmath}
\usepackage{amssymb}
\usepackage{booktabs}
\usepackage{multirow}
\usepackage{graphicx}
\usepackage{subcaption}
\usepackage{pgf}
\usepackage{tikz}
\usetikzlibrary{arrows.meta,calc,positioning,fit,backgrounds,shapes.misc}
\usepackage{xcolor}
\usepackage{url} %

\newcommand{\tup}[1]{\langle #1 \rangle}

\usepackage[preprint]{corl_2026}

\title{REAP: Learning General Rearrangement Heuristics from Motion-Planner Feedback}

\author{
  Daniel Swoboda \qquad Hector Geffner\\
  Chair of Machine Learning and Reasoning\\
  RWTH Aachen University, Germany\\
  \texttt{\{daniel.swoboda, hector.geffner\}@ml.rwth-aachen.de}
}

\begin{document}
\maketitle

\begin{abstract}
    Robotic rearrangement tasks require planning sequences of actions that
    fulfill high-level specifications while also being geometrically feasible.
    Symbolic abstractions make the combinatorial structure of rearrangement
    tasks tractable, but they usually do not fully capture the geometric
    constraints that determine which actions a robot can actually realize.
    These constraints are hard to express in the symbolic model.
    We propose \textbf{REAP} (Rearrangement with Elicited Action Preconditions), a framework in which a
    relational graph neural network heuristic is learned over a simple
    symbolic abstraction. A motion planner serves as a supervisor that elicits
    the \emph{hidden geometric preconditions} of the problem during training.
    Deployed with greedy best-first search, the learned heuristic generalizes
    out-of-distribution along object count, plan length, and object arrangement, and can be trained jointly on several domains.
    Compared to classical heuristic search and learning-based baselines, REAP solves
    more problems, requires roughly an order of magnitude fewer motion-planner queries
    than comparable methods, and the action it ranks first is geometrically feasible in 94\% of
    sampled states, against 22\% for a random choice. 
    
    \textbf{Website:} \url{https://elicited-preconditions.github.io}.
\end{abstract}

\keywords{Task and Motion Planning, Generalized Planning, RL}

\par\noindent\begin{minipage}{\textwidth}
    \centering
    \includegraphics[width=1\textwidth]{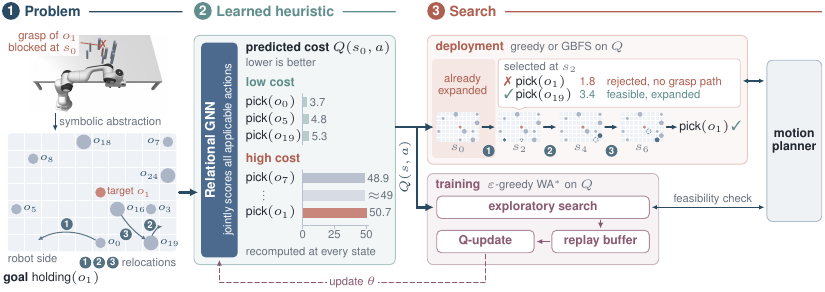}
\captionof{figure}{
REAP learns which symbolically applicable actions are geometrically feasible, without
modeling the geometric constraints.
\textbf{(1)}~Grasping the target $o_1$ is symbolically applicable, but not geometrically
feasible since other objects block the target. \textbf{(2)}~A relational GNN, whose input
state includes the goal, predicts a cost $Q(s,a)$ for each applicable action and applies to
any number of objects. Feasible picks of blocking objects get low costs, infeasible picks high
costs. \textbf{(3)}~At test time, a plan is found by greedy execution or greedy best-first
search on $Q$, with the motion planner checking each expanded action. $Q$ is learned by an RL
algorithm that uses $\varepsilon$-greedy WA$^\star$ to guide the exploration and penalizes
infeasible actions.}
\label{fig:method-overview}
\end{minipage}

\section{Introduction}

Robots performing tasks in cluttered environments like clearing a table,
retrieving an occluded target, or constructing an object from parts must
select \emph{what} to do symbolically and \emph{how} to realize each step
as a collision-free motion. These tasks are examples of combined
task and motion planning (TAMP)~\citep{garrett2020pddlstream,garrett2021integrated},
where a symbolic plan reaching the given goal is sought and every action
of that plan must be geometrically feasible. The simplest approach
incrementally searches for plans while checking the feasibility of each
candidate with a motion planner. This, however, is too inefficient as
the geometric constraints play only a passive role in the
search~\citep{kaelbling2011hierarchical,srivastava2014combined}. Other
approaches make the geometric constraints explicit in the planning
formulation, but modeling such constraints and solving the resulting
joint problem is similarly
hard~\citep{toussaint2015logic,kingston2018sampling}. A third line of work
aims to amend the heuristics used to guide the
search to take into account both the goals and the geometric
constraints. The resulting search is more informed, and the challenge
is the formulation and computation of such hybrid
heuristics~\citep{wells2019learning,driess2020deep,khodeir2023streams,yang2023sequence}.

In this work, we propose a novel approach for addressing robot rearrangement tasks
based on heuristic estimators leveraging recent progress on generalized planning,
where policies are learned for solving large collections of classical planning problems
\citep{martin-geffner-2000,boutilier-et-al-ijcai2001,srivastava-et-al-aaai2008,jimenez2019review}.
Some of these methods learn these policies in terms of parametric value functions $Q_\theta(s,a)$
represented in relational graph neural networks (R-GNN) and trained without supervision using reinforcement learning
\citep{stahlberg2023learning,stahlberg2026first}. As in goal-conditioned RL
\citep{andrychowicz-et-al-nips2017,nasiriany-et-al-neurips2019,eysenbach-et-al-neurips2022},
the atoms in $s$ represent both the current state and the goal of the instance.

Generalized symbolic planning policies are particularly relevant in this setting in two ways.
On the one hand, a single policy generalizes across instances varying in object count, initial
state, and goal by acting greedily on the learned $Q$-value function.
On the other, the policy can be learned \emph{model-free}: it does not need to know the action
preconditions and effects a priori.
The training loop only requires a simulator that, for each state $s$, exposes the applicable
action set $A(s)$, the successor $s' = f(a,s)$, an action cost $c(a,s)$, and a goal check.\footnote{Even a 
uniform-random policy needs these signals: knowing which actions are applicable, advancing the state, and detecting the goal is the minimum any policy needs to act. }
What distinguishes TAMP from classical planning, given a suitable symbolic abstraction,
is that not every symbolically applicable action is \emph{geometrically executable}.
Alongside $A(s)$ there is a hidden function $X(s) \subseteq A(s)$ encoding the set of
\emph{feasible} actions, decidable by the motion planner provided $s$ encodes a suitable
abstraction of the geometry. We illustrate this in Figure~\ref{fig:method-overview} on one
of our rearrangement tasks where typically $\sim$80\% of the symbolically applicable actions are not
executable. 
Making the symbolic policy aware of this hidden function is a minor modification to the
RL algorithm: selected actions $a \in A(s)$ found not to be executable in $s$, i.e.,
$a \in A(s) \setminus X(s)$, are penalized in the state $s$. The learned $Q$-values
then reflect both the number of steps to the goal and the feasibility of the actions.

In order to have a complete TAMP planner, REAP (Rearrangement with Elicited Action Preconditions)
\emph{uses} the learned $Q$-values not to push greedily for the goal, but to guide a
greedy best-first search \citep{agostinelli2024q,aichmueller2026learning}, over a coarse 
abstraction of the geometry, that uses grid-based adjacency relations and occupancy.
The learned heuristic solves more problems than baselines, generalizes out-of-distribution
along object count, plan length, and arrangement, and requires an
order of magnitude fewer motion-planner queries at run time than classical
heuristic search and learning-based baselines. Removing the motion-planner feedback
collapses the framework to symbolic search, like our BrFS and
$h_{\text{FF}}$ baselines, but guided by the learned $Q_\theta(s,a)$ values. We
demonstrate this approach across five rearrangement environments with
distinct hidden-precondition structures, using the same architecture and training loop
across all of them.

We contribute REAP, a TAMP framework that (i) trains a generalized TAMP planning heuristic with
motion-planner verdicts as the RL signal, (ii) requires an order of magnitude
fewer motion-planner queries at deployment, (iii) scales without retraining to
OOD instances, and (iv) shows that the same approach applies across different robotic rearrangement
environments.

\section{Related Work}
\label{sec:related}

Integrating planning, learning, and execution is a long-standing problem in robotics, 
where a symbolic plan is only useful insofar as it grounds out in physically executable 
motion. Prior work approaches this from several directions, which we group below.
Related work on \textbf{generalized planning} was discussed in the introduction.

\paragraph{TAMP with Learned Components.}
Classical task-and-motion planning couples symbolic search with parameter
sampling and feasibility checks in an integrated search loop~\citep{kaelbling2011hierarchical,garrett2020pddlstream,garrett2021integrated}.
Various works attempt to replace parts of this stack with learned components.
The closest to ours is \citet{khodeir2023streams}, a GNN heuristic guiding
search in a stream-based TAMP planner~\citep{garrett2020pddlstream}.
They use offline learning from past execution traces to prioritize the sampling
of numerical parameters for the planners (modeled as streams). 
Another predecessor of our work is \citet{wells2019learning}. They learn an SVM-based 
binary feasibility classifier from offline trajectories with motion planner feedback, and use it
to prune candidate task plans before invoking the motion planner.
Other supervised approaches predict action feasibility from visual observations~\citep{driess2020deep}, 
or use diffusion models as learned samplers for the continuous parameters of a TAMP solver~\citep{fang2023dimsam}. 
\citet{yang2023sequence} 
move from single-action feasibility classifiers to predicting feasibility for an 
entire candidate plan skeleton. \citet{curtis2024proc3s}~use 
LLMs to generate continuous CSPs from the task description to constrain the search space.
Closest to REAP is \citet{kim2019learning}, who also learn a relational value function with a GNN to guide
a TAMP planner. Their approach (SAHS) differs in where geometric feasibility enters.
SAHS encodes it in the state representation, through crafted predicates defined over swept
volumes that require inverse kinematics and motion planning to evaluate. We 
learn a heuristic to guide the planning directly, leaving feasibility out of the symbolic 
representation entirely and learning to avoid infeasible actions during training. 
Their target
$Q$-values are derived from successful plans produced by an oracle planner and scale
with the number of remaining steps, so their training data contains no dead-ends and no
infeasible actions. REAP's training data contains both.
\citet{silver2021planning} (PLOI) rank object
importance from demonstrations in order to plan on a problem with a reduced number of objects, which is
efficient when few objects matter but degrades when most objects are genuine occluders.
\citet{cieslar2025learning} predict the refinability of a given plan skeleton to guide
continuous parameter search. We assume no such skeleton, but instead learn a policy that
guides the planning. Furthermore, we are agnostic to the controller, assuming we only need
to select a high-level action, and not how to realize it. 
\citet{chitnis2022learning} learn symbolic operators from demonstrations
for both high-level planning and low-level sampling, whereas we learn action values
over pre-defined operators without direct access to low-level information.

\paragraph{Object Rearrangement.}
REAP's evaluations adapt environments from prior rearrangement
benchmarks: confined-shelf rearrangement~\citep{wang2022lazy},
multi-layer shelf retrieval~\citep{bouhsain2025longhorizon}, and multi-level
stacking~\citep{kulshrestha2023scl}. Adjacent work on non-prehensile
manipulation among movable obstacles~\citep{saxena2023em4m} shares our
cluttered, geometrically constrained setting. REAP focuses on
prehensile actions.

\paragraph{Other Learning Approaches in TAMP.}
SPOT~\citep{saha2025spot} plans without symbolic discretization, running
A* directly over a sequence of transformations of point-cloud scenes. 
\citet{shah2025lamp} learn
relational predicates and PDDL-style operators from unsegmented motion
demonstrations, then plan symbolically on instances with longer
horizons and far more objects than demonstrated.
Further afield, \citet{dhakal2026anticipatory} learn to weight multiple
candidate plans, based on the incurred costs of future tasks in the same environment.
\citet{jung2025spin} distill a motion-planning algorithm into a reactive
policy via imitation, eliminating search at deployment. End-to-end
visuomotor policies~\citep{chi2023diffusion} produce motion trajectories directly
without an explicit planning step.

\section{Method}
\label{sec:method}

REAP learns general
 heuristics for rearrangement problems with hidden
geometric preconditions. These hidden preconditions are not learned explicitly in syntactic form but implicitly
in a $Q$-value function that guides the search for feasible plans. The framework consists of three main components:
(i) a simplified symbolic abstraction of the rearrangement problem in the form of a classical planning domain,
(ii) a relational graph neural network heuristic trained via reinforcement learning and
feedback from a feasibility oracle that results in a learned $Q$-value function, and (iii)
a greedy best-first search procedure that uses the learned $Q$-function
as a heuristic to guide the search for a feasible plan. The learned $Q$-value anticipates the
geometric feasibility of actions in the symbolic planning state that must
provide a suitable abstraction of the configuration space. As a result,
the search for plans that are valid in the explicit planning abstraction
while complying with the implicit geometric constraints is informed
by the two dimensions of the combined task and motion problem.
We describe the components of REAP below. Figure~\ref{fig:method-overview} visualizes the
three components on an open-tabletop instance.

\subsection{Generalized Policies with Hidden Geometrical Preconditions}
\label{sec:method:setup}

The \emph{symbolic abstraction} is expressed as a PDDL/STRIPS planning problem $P=\tup{D,I}$
where $D = \tup{Pred, Act}$ is a planning domain with predicates $Pred$ and action schemas $Act$,
and $I$ is the instance information $I=\tup{O,Init,G}$ where 
$O$ is a set of objects (constants), while $Init$ and $G$
are sets of ground atoms formed from the domain predicates and the objects,
the first representing the initial state of the problem $P$, and $G$ representing
the goal condition. Coupled with the PDDL specification is a simulation environment
with a feasibility oracle for every symbolic action. In our environments, 
the oracle combines inverse kinematics with either online motion planning or a library of pre-computed
collision-checked motions.

 The world is modeled as a set of fixed and movable objects. The latter can be moved
 across one or multiple grids (2D or 3D) with the help of a fixed robot arm.
 The predicate set $Pred$ models the discrete physical relations: occupancy of cells or 
 support surfaces, gripper state, and support relations between objects. The action 
 schemas $Act$ include actions of the form \texttt{pick(obj)} and \texttt{place(obj, target)}
 or suitable variations. The ground actions in an instance are obtained by instantiating the
 action schemas with the objects. A ground action $a$ is applicable in state $s$,
 written $a \in A(s)$, iff $\textsc{pre}(a) \subseteq s$, and yields the successor
 $s' = f(a,s) = (s \setminus \textsc{del}(a)) \cup \textsc{add}(a)$. A plan for $P$
 is a sequence $a_0,\ldots,a_n$ with $a_i \in A(s_i)$, $s_0=Init$, and $G \subseteq s_{n+1}$.
 Here $\textsc{pre}(a)$, $\textsc{del}(a)$ and $\textsc{add}(a)$ are the standard
 STRIPS precondition, delete and add lists of the action schema.
 
 In addition to the symbolic abstraction $P=\tup{D,I}$, our framework assumes that there is a hidden function $X(s)$ that represents the geometrically feasible actions
 in $s$. We check whether an applicable action $a \in A(s)$ is (geometrically) feasible; namely, $a \in X(s)$, by calling a motion planner. This assumes
 that the symbolic state $s$ provides a suitable abstraction of the space. The plan $a_0, \ldots, a_n$ above for $P$ is then feasible if
 each action $a_i$ in the plan is feasible in state $s_i$ where it is executed, namely $a_i \in X(s_i)$.
 The symbolic abstraction is intentionally a coarse over-approximation of geometric
reality: we model only cell occupancy and, in 3D settings, support relations. A
symbolically applicable action may therefore have no collision-free trajectory
realizing it (e.g., due to joint limits or obstacles blocking the path). The motion-planner
check $a \in X(s)$ is the expensive operation, and REAP is designed to minimize
the number of such calls before a feasible plan is found.

\subsection{Relational GNN Trained with RL}
\label{sec:method:gnn}
The parametric $Q_\theta$ function maps state-action pairs $(a,s)$
into real values $Q_\theta(s,a)=Q(s,a)$. As in reinforcement learning, $Q(s,a)$ is the
estimated cost of taking action $a$ in state $s$ and following the policy thereafter;
here that cost is the number of remaining actions to reach the goal. Equivalently, $Q_\theta$
is the negated action value of standard $Q$-learning with a reward of $-1$ per step.
Following \citet{stahlberg2022learning,stahlberg2023learning}, 
the function is represented as a relational-GNN that takes states $s$ as input, extended
with atoms representing the goals and the applicable actions $a$ in $s$, and
outputs all $Q_\theta(s,a)$ values in a single pass over the R-GNN \citep{stahlberg2026first}.
The weights of the GNN are learned via a deep-Q-learning algorithm \citep{mnih-et-al-nature2015}
but with the greedy rollouts that accommodate different forms of exploration replaced by a
systematic best-first search: weighted A* with evaluation function $F(s,a)=g(s)+ w Q_{\theta}(s,a)$,
where $g(s)$ is the accumulated cost to reach $s$, and $Q_{\theta}(s,a)$ is the cost estimate of
reaching the goal from $s$ by starting with the action $a$.
Since the learned $Q$-values are not admissible lower bounds, we add an
exploration component to WA*: with probability $\varepsilon$, 
a newly generated node is moved to the front of the OPEN list 
and is expanded next regardless of its score, and expanded
states are placed in a replay buffer for $Q$-value updates~\citep{aichmueller2026learning}.
Encoding states and goals as atoms enables generalization to new instances;
systematic exploration enables sparse goals and long
plans~\citep{agostinelli-et-al-nmi2019,orseau-lelis-aaai2021}.
Since a single pass through the R-GNN yields the $Q_\theta(s,a)$ for all the applicable
actions $a$ in a state, the nodes in the search do not represent states $s$, but state-action
pairs $(s,a)$, which implicitly encode the successor states $s'=f(a,s)$ \citep{agostinelli2024q}.

During \emph{training}, $w=2$ in the WA* search, while at \emph{test time},
a greedy best-first search is used instead that discards the
accumulated costs $g(s)$ and only uses the $Q_{\theta}(s,a)$ estimates.
The search expands the $(s,a)$ pair in the priority queue with the lowest evaluation
by applying the action $a$ in $s$ and obtaining the resulting state $s'$
and the $Q_{\theta}(s',a')$ values for the actions $a'$ applicable in $s'$.
Before this expansion, we check if $s$ is a goal state, and if not, if the action
$a$ is executable in $s$, i.e., if $a \in X(s)$. 
Feasibility verdicts are cached, so a violation at one $(s,a)$ is reused on
subsequent visits. 
During training, the infeasible action that reaches a state's retry limit receives the dead-end target.

At test time, we also evaluate a purely greedy execution variant, \textsc{REAP-greedy}, that commits to
$\arg\min_a Q_\theta(s,a)$ at each step without backtracking. Since the motion planner is stochastic, we allow
up to $2$ retries per state. Problems solved by \textsc{REAP-greedy} are
thus solved without search.
In both variants every action is verified by the feasibility oracle in the state
where it is expanded, so the learned heuristic determines only \emph{which} actions are
checked first, never whether an unverified action can enter a plan.
\textsc{REAP-GBFS}, on the other hand, backtracks and can recover from errors in the
learned heuristic. It is probabilistically complete on the finite symbolic state space,
given a probabilistically complete motion planner. In practice, finite planner budgets, precomputed motions and timeouts
break the completeness guarantee. 

The same R-GNN architecture is used across all five environments in
Section~\ref{sec:result:setup}, each with its own set of predicates, actions, and
object types; layer count, embedding size, and activation are detailed in
Appendix~\ref{sec:appendix:training-details}.

\section{Experiments}
\label{sec:result}

We evaluate REAP on five rearrangement environments with different
hidden-precondition and task structures, and compare against classical search,
the widely used TAMP planner PDDLStream and three learning-based methods.
The learned heuristic generalizes out-of-distribution along several axes, and the same architecture 
and training loop are used in all five environments.

\subsection{Setup}
\label{sec:result:setup}

\paragraph{Simulator and Robot.} 
All experiments use a single Franka Emika Panda arm in MuJoCo~\citep{todorov2012mujoco}, with 
inverse kinematics solved by MINK~\citep{zakka2024mink} or analytically~\citep{he2021franka}.
Motions are planned with RRT-Connect~\citep{kuffner2000rrtconnect}, partly via VAMP~\citep{thomason2024vamp},
online or precomputed in a motion library.
The PDDL domains are simple grid abstractions; scenes and feasibility checks are adapted to each environment.

\begin{figure*}[!t]
  \centering
  \includegraphics[trim={300 100 200 200}, clip, width=0.19\linewidth]{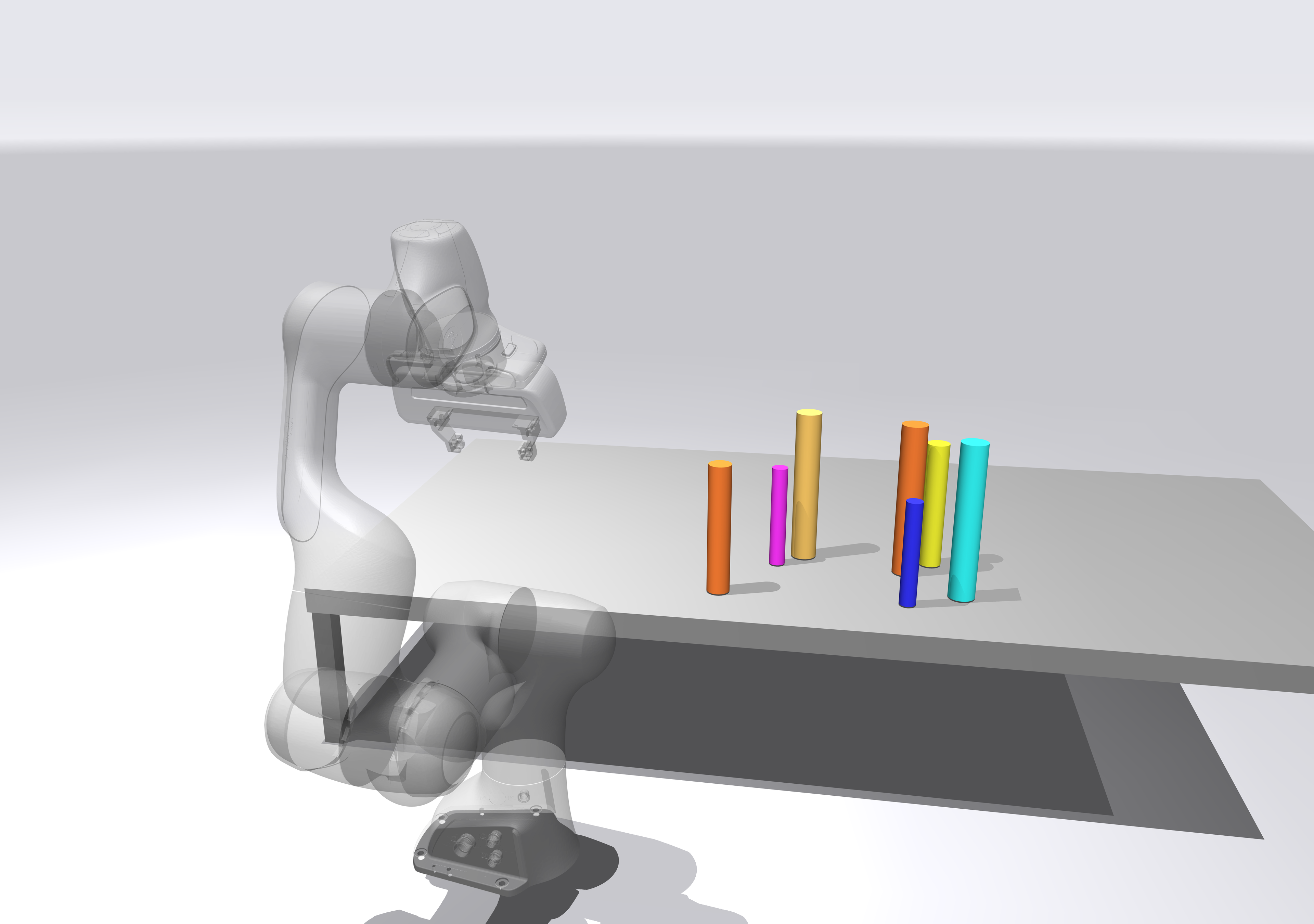}\hfill
  \includegraphics[trim={300 100 200 200}, clip, width=0.19\linewidth]{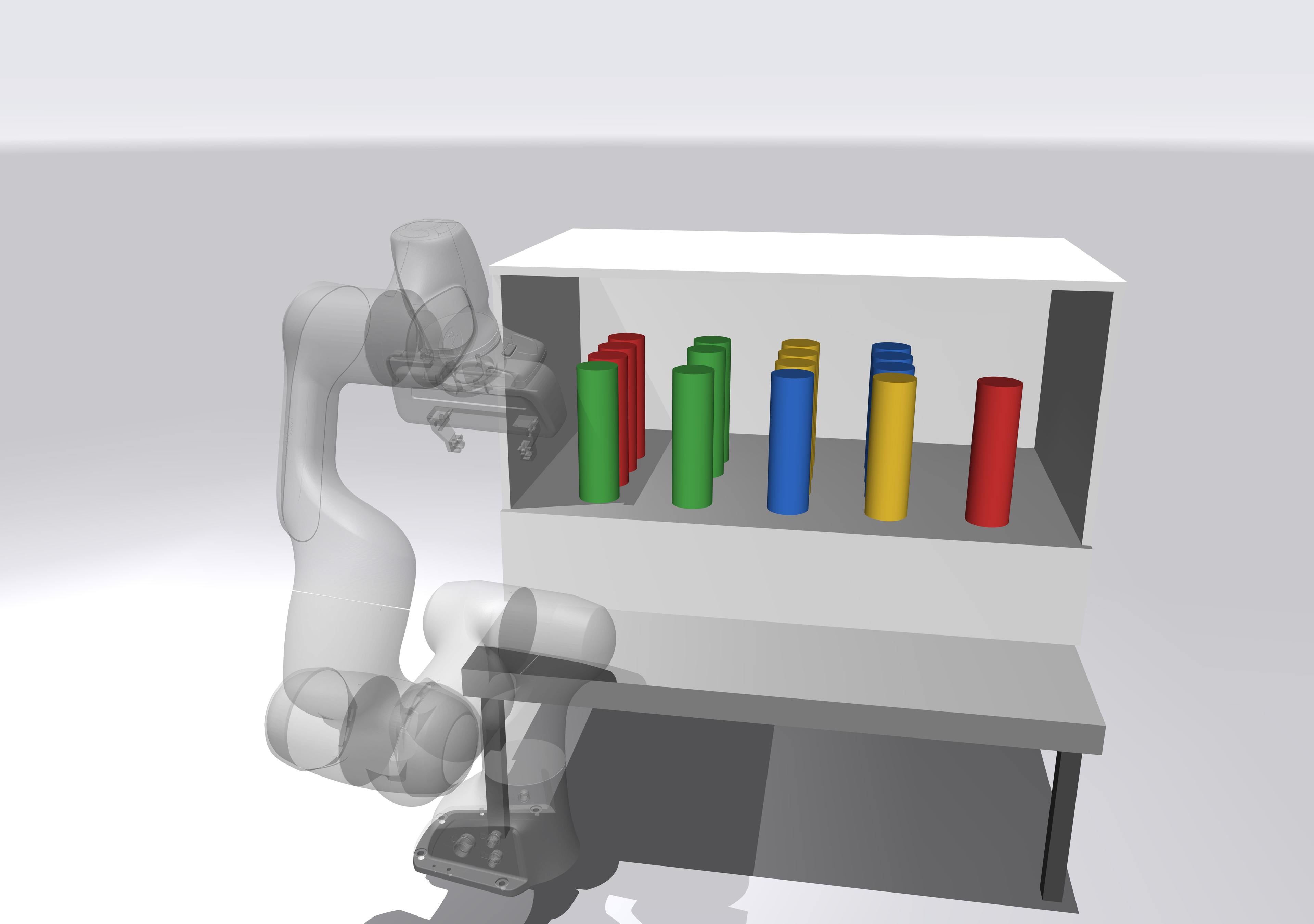}\hfill
  \includegraphics[trim={456 180 200 240}, clip, width=0.19\linewidth]{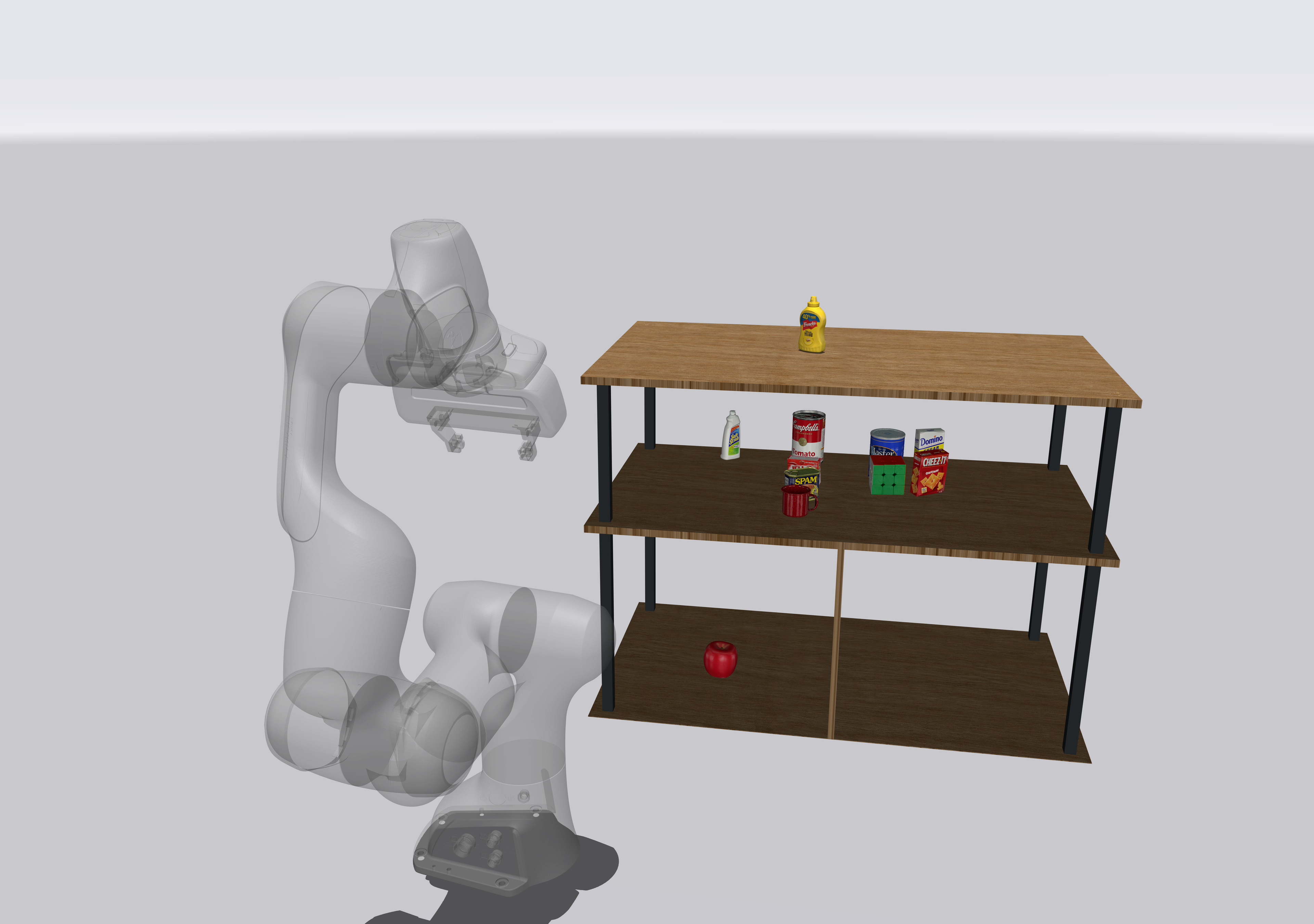}\hfill
  \includegraphics[trim={300 100 200 200}, clip, width=0.19\linewidth]{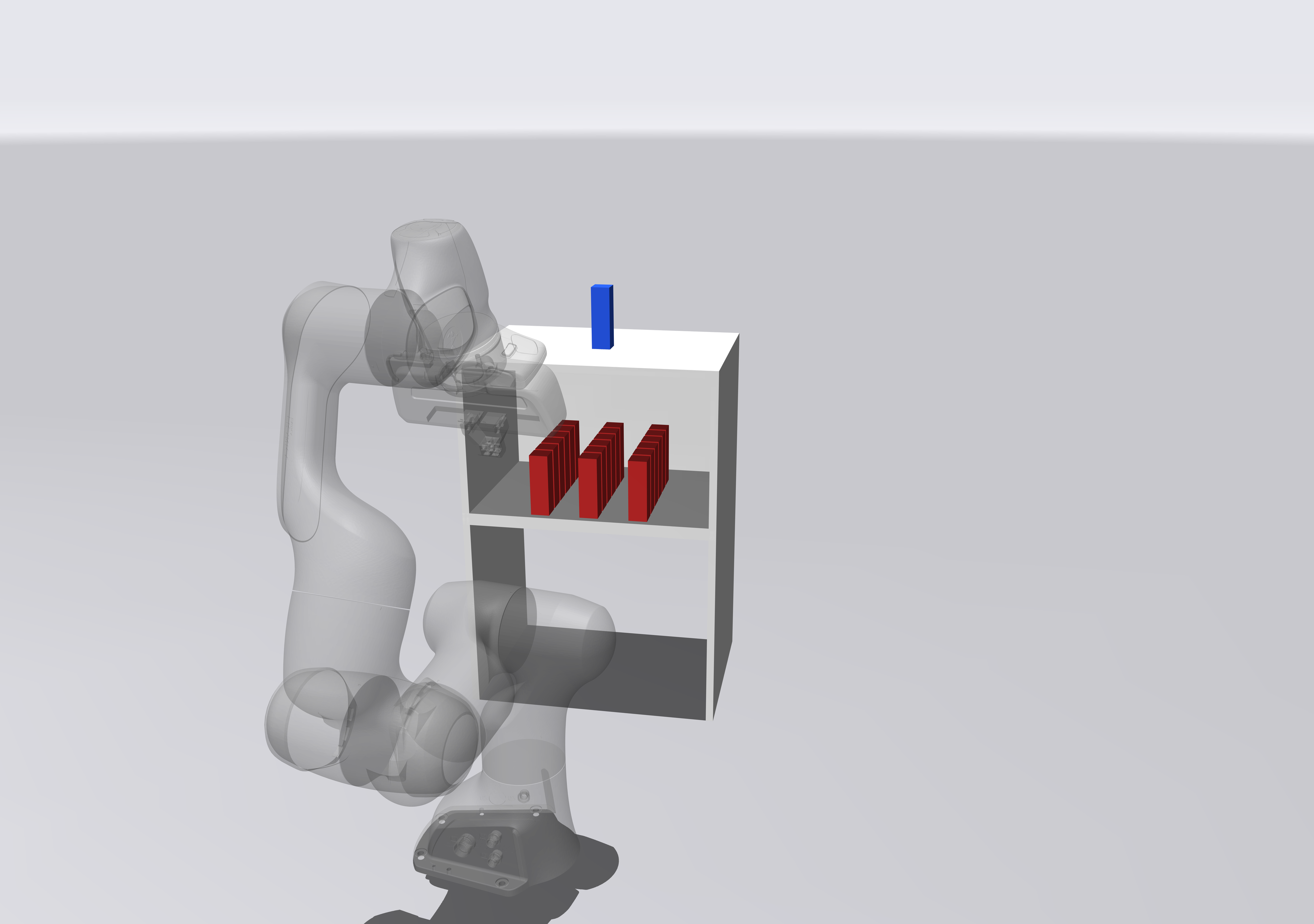}\hfill
  \includegraphics[trim={300 100 200 200}, clip, width=0.19\linewidth]{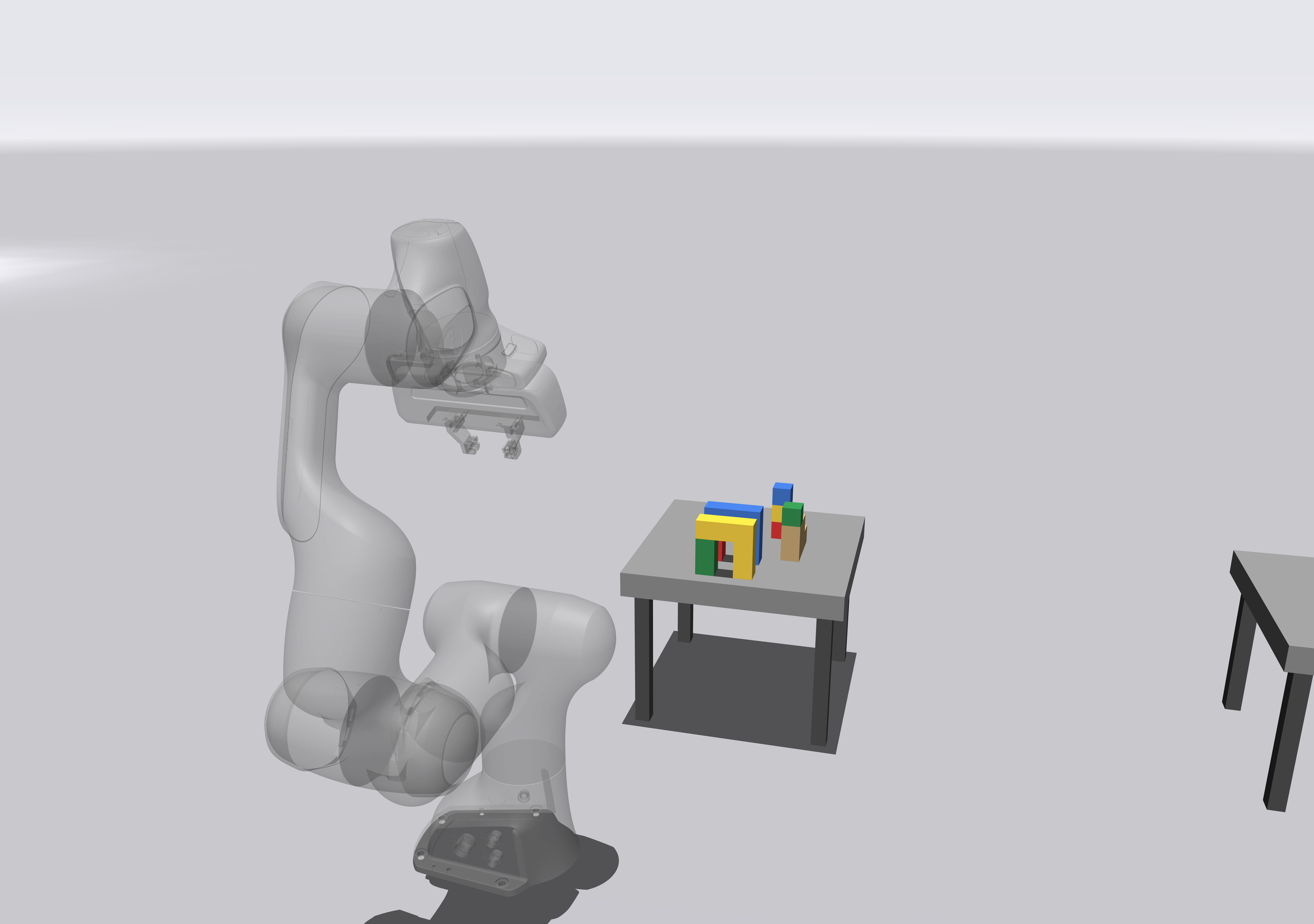}
  \\[0.3em]
  \makebox[0.19\linewidth]{(a) open-tabletop}\hfill%
  \makebox[0.19\linewidth]{(b) confined-shelf}\hfill%
  \makebox[0.19\linewidth]{(c) multi-level shelf}\hfill%
  \makebox[0.19\linewidth]{(d) Access-19}\hfill%
  \makebox[0.19\linewidth]{(e) 3D stacking}%
  \caption{REAP instantiates the same relational-GNN architecture
  trained with motion-planner feedback over hidden geometric preconditions
  across five evaluation environments, each with a different hidden geometric precondition.
  Gripper clearance on the open table, lateral grasp clearance in the confined shelf, blocking in a multi-level
shelf, front-facing retrieval past blockers (Access-19), and finger/arm interference in
3D stacking. At the same time, REAP also learns the cost-to-go of the underlying planning problems, to
guide search towards efficient solutions.}
  \label{fig:teaser}
\end{figure*}

\paragraph{Environments.} Each environment is specified by a PDDL domain, a
simulation environment, and a problem generator producing solvable instances. %
The hidden geometric preconditions are implicit, following the task layout, the environment geometry, and the robot's kinematics.
The evaluation is carried out on five environments.
The first is an \emph{open-tabletop rearrangement} setting that serves as
the primary evaluation domain: a 10$\times$7 grid of cells on a table
with cylinders occupying single cells, focused on object rearrangement, depicted in Figure~\ref{fig:teaser}a.
The other four environments test how our approach applies across different rearrangement settings.

\paragraph{Baselines.} In each domain, the
same symbolic action space, feasibility-check interface, and
simulation backend are used in all methods to ensure a clean comparison. We compare against BrFS, breadth-first search over the 
symbolic action space, with the MP filter applied at every expansion, and
$h_{\text{FF}}$, greedy best-first search ordered by the domain-independent
$h_{\text{FF}}$-heuristic~\citep{hoffmann2001ff}.
These isolate the contribution of the \emph{learned} value heuristic: 
BrFS holds learning and heuristic-design constant at zero, 
$h_{\text{FF}}$ holds learning at zero while introducing a classical
domain-independent heuristic. PDDLStream~\citep{garrett2020pddlstream}
is a widely used TAMP planner that interleaves symbolic search 
with sampling of continuous action parameters.
Four configurations of PDDLStream are tested, varying stream definitions and search
strategies, including a combination of an informed pick and place generator with a symbolic occlusion
axiom in \texttt{pick}'s precondition.
We further compare against three learning-based methods, referred to below by the
names: SAHS~\citep{kim2019learning,kim2022representation}, which learns a relational
value function over a geometrically enriched state representation.
PLOI~\citep{silver2021planning}, which ranks object importance in order to plan on a
reduced problem. And BT-h~\citep{cieslar2025learning}, which predicts the refinability
of a given plan skeleton. We reimplement each against REAP's symbolic action space and
feasibility-check interface. SAHS and PLOI are trained on REAP's training problems and receive
only the initial state and goal, as REAP does. For SAHS, every empty grid cell is a
candidate placement region. PLOI uses its
paper's architecture and loss, with our planners in place of Fast Downward. BT-h is
additionally given the object ordering of a reference plan (found by our generator's feasibility-checked search) 
and refines the placement cells only. 

\subsection{Benchmark Results}
\label{sec:result:headline}

\begin{table*}[!t]
\centering
\caption{Results on $300$ open-tabletop problems, all methods under a
$300$\,s per-problem cap. \emph{MP~/~plan} is the mean number of distinct
$(\text{state},\text{action})$ motion-planner queries per solved problem, median in brackets. 
\emph{Exp./steps} is search
expansions over plan length.
\emph{Checks/exp.} is motion-planner queries per expansion, a higher number indicates more infeasible queries. Time is averaged
over \textit{solved} problems. \textsc{REAP} rows are mean $\pm$ std
over $5$ seeds. All baseline rows are single runs.}
\label{tab:opentabletop}
\small
\resizebox{\textwidth}{!}{%
\begin{tabular}{lccccc}
\toprule
Method & Solve (\%) & MP / plan [med.] & Checks/exp. & Exp./steps [med.] & Time (s/solved) \\
\midrule
\textsc{REAP-GBFS}   & \textbf{95.9}\,$\pm$\,0.5 & \textbf{17.1}\,$\pm$\,4.2 [3.1] & \textbf{1.96} & \textbf{1.29} [1.00] & \textbf{3.9} \\
\textsc{REAP-greedy} & 89.5\,$\pm$\,1.7 & 5.2\,$\pm$\,0.7 [3.0] & n/a & n/a & 1.7 \\
\midrule
BrFS                 & 56.7 & 148.8 [12.0] & 7.01 & 7.26 [3.00] & 19.5 \\
$h_{\text{FF}}$      & 76.7 & 179.0 [1.0]  & 5.04 & 3.62 [1.00] & 25.4 \\
\midrule
SAHS~\citep{kim2019learning,kim2022representation} & 51.0 & 2937.4 [2857.0] & n/a & 1.37 [1.00] & 68.8 \\
PLOI\,+\,$h_{\text{FF}}$~\citep{silver2021planning} & 72.3 & 210.7 [2.0] & 4.15 & 8.26 [1.00] & 36.2 \\
PLOI\,+\,BrFS~\citep{silver2021planning}          & 53.3 & 97.3 [2.0] & 3.18 & 10.70 [1.00] & 18.0 \\
BT-h~\citep{cieslar2025learning}, \emph{given the plan ordering} & 81.7 & 28.5 [3.0] & n/a & n/a & 4.5 \\
\midrule
PDDLStream (test-stream, ff-astar2)   & 41.7 & 1.7 [1.0]  & n/a$^{a}$ & n/a$^{a}$ & 5.2 \\
PDDLStream (test-stream, ff-eager)    & 42.7 & 2.1 [1.0]  & n/a$^{a}$ & n/a$^{a}$ & 2.8 \\
PDDLStream (cell-generator, ff-eager) & 43.7 & 9.8 [1.0]  & n/a$^{a}$ & n/a$^{a}$ & 11.5 \\
PDDLStream (informed cells + occlusion axiom) & 64.3 & 23.6 [1.0] & n/a$^{a}$ & n/a$^{a}$ & 17.2 \\
\bottomrule

{\small\raggedright $^{a}$ PDDLStream exposes no expansion counter.}
\end{tabular}}

\end{table*}

All baseline methods and REAP are evaluated on the same 300-problem open-tabletop suite (Table~\ref{tab:opentabletop})
consisting of problems with 10--24 objects and reference plan lengths from 1 to 15, all verified to be solvable
with the motion planner.
\textsc{REAP-GBFS} solves more problems than both classical baselines ($95.9\%$ vs.\ $76.7\%$
and $56.7\%$) with fewer expansions per plan step. 
Its median search-overhead ratio is $1.00$: on more than half of solved problems 
\textsc{REAP-GBFS} expands only states that lie on the final plan, 
reaching the goal without backtracking. The gap widens on problems that
require long action sequences, where BrFS and $h_{\text{FF}}$ largely time out.
Because BrFS and $h_{\text{FF}}$ share REAP's symbolic action space and
motion-planner filter, the comparison isolates the contribution of the
\emph{learned} value heuristic.

 The integrated task-and-motion-planning baseline, PDDLStream, solves $\sim$$98\%$ of instances that require no
 relocation, but its solve rate collapses to $\leq22\%$ once one object must be relocated
 in the three standard configurations, and to zero beyond. Including an occlusion axiom and informed cell
 placements provides more signal to the search and raises the solve rate to $91\%$ with one relocation
 and $34\%$ with two or three, but none beyond that. 
 Across the learning-based baselines, BT-h is the strongest, but is given a reference plan. SAHS and PLOI
 degrade on longer plans and harder arrangements (Fig.~\ref{fig:ood}), and use significantly
 more motion-planner queries per solved problem than REAP.

Across the initial states of the $300$ problems, $80\%$ of
symbolically applicable actions are geometrically infeasible.
The action ranked first by REAP's $Q_\theta$ is infeasible in only
$1.7 \pm 0.6\%$. On $1{,}327$ states reached by short random walks, it is infeasible in $5.9 \pm 0.7\%$, wheras random choice is infeasible in $78\%$.

\subsubsection{Out-of-Distribution Generalization}
\label{sec:result:ood}
\begin{figure}[!t]
  \centering
  \includegraphics{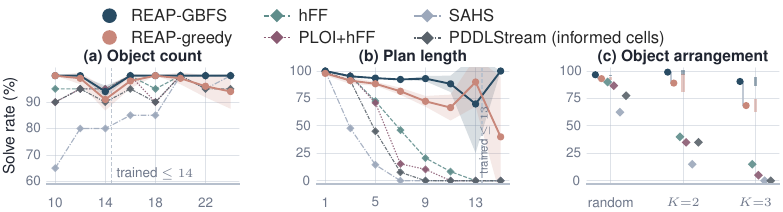}
  \caption{Out-of-distribution generalization along three axes on the open-tabletop
  environment, for \textsc{REAP} and four baselines. Each policy is trained on
  problems with up to $14$ objects, reference plan lengths up to $13$, 
  and random scene placement, and is
  evaluated on object counts up to $24$ (a) and plan lengths up to $15$ (b). (c)
  covers arrangements forcing at least $K$ direct blockers of the OoI to be removed.
  Vertical dashed lines mark the training-distribution boundary. 
  }
  \label{fig:ood}
\end{figure}

We evaluate the same value heuristic against three
out-of-distribution axes relative to the training set:
\emph{object count}, \emph{plan length}, and \emph{object arrangement}.
On open-tabletop rearrangement, the policy trained on at most $14$ objects
with plan lengths up to $13$ reaches $95.9 \pm 0.5\%$ solve rate under
\textsc{REAP-GBFS} across the full evaluation suite of $300$ verified
problems within a $300$\,s per-problem compute cap. The solved subset includes
up to $24$ objects and plan lengths up to $15$ steps. Figure~\ref{fig:ood} shows
that \textsc{REAP-GBFS} stays at or above $88.3\%$ on every object-count,
plan-length, and object-arrangement bucket that contains more than two problems
per seed, while the
greedy variant degrades on longer plans and harder arrangements.

\subsection{Beyond Open-Tabletop Rearrangement}
\label{sec:result:perenv}

We apply REAP with few changes to four further environments,
each adapted from a published robot manipulation benchmark.
Across all four environments REAP keeps the same architecture, training loop, 
and motion-planner-as-supervisor scheme. 
The PDDL domain, scene generator and hyperparameters are adapted per environment,
and the supervising motion-planning stack to its geometry.

\paragraph{Confined-Shelf Rearrangement (Fig.~\ref{fig:teaser}b).}
\citet{wang2022lazy} address grid-based confined-shelf rearrangement
where same-colored cylinders must share a column, solved with a
classical lazy-evaluation planner under hand-engineered reachability
constraints. We re-implement the task as a PDDL domain over a
$9\times4$ grid ($34$ reachable cells) with up to $16$ cylinders in up to four color groups, leaving
lateral grasp reachability to the motion planner. REAP is trained
only on monotone instances ($3$--$9$ objects) and evaluated at the
same complexity range as theirs: monotone (in-distribution, $3$--$9$
objects) and non-monotone (OOD, $10$--$16$ objects) configurations, the latter
requiring an intermediate parking move. Our results (Table~\ref{tab:env:summary})
show that REAP generalizes to the OOD tasks, achieving 100\% coverage at $16$ objects.

\paragraph{Multi-Level Shelf Retrieval and Access-19 (Fig.~\ref{fig:teaser}c,d).}
\citet{bouhsain2025longhorizon} define the \emph{Access} problem: retrieve
an object surrounded by blockers across a multi-layer shelf. They refine it in
the \emph{Access-19} instance with $18$ blockers arranged in a grid.
Their approach predicts feasibility from continuous
poses. We re-implement both tasks as distinct PDDL domains, but restricted to front-facing
grasps. For multi-level retrieval, REAP is trained on $1$--$5$
blockers (mean $3$) and evaluated OOD on $4$--$10$ (mean $7$), which is similar
to the original but with lower object densities.

A single goal-conditioned policy solves Access-19 in distribution (up to $14$
blockers, Table~\ref{tab:env:summary}) but none of the $18$-blocker benchmark instances,
and only $38.9 \pm 16.1\%$ of $72$ unseen $18$-blocker layouts.
Splitting the task into two policies, one that
extracts the target and one that restores the initial state for the blockers,
executed in sequence, solves
$89.8 \pm 2.9\%$ of the $18$-blocker layouts. The
$18$-blocker layouts appear to require a decomposition into subproblems, which our single policy does not learn.

\paragraph{Three-Dimensional Block Stacking (Fig.~\ref{fig:teaser}e).}
\citet{kulshrestha2023scl} infer object-dependency graphs to plan
multi-level block stacking, evaluated only via pose-matching success. 
Their problem does not transfer well to a symbolic setting, but the same task
structure, building a target structure from a pile of blocks, was the inspiration for our
environment. It tests REAP's generalization to $3$D grids.
Objects are placed unassembled, goals represent fully assembled multi-part structures.
Multiple symbolic pick and place actions capture object types and $3$D rotations
within the $3$D grid.  REAP is
trained on structures with  up to $8$ blocks and evaluated OOD on
combined structures of $4$--$12$ blocks. Table~\ref{tab:env:summary} shows that OOD
overhead is roughly twice that of the $2$D cases, while success rate remains high.

\begin{table}[!t]
\centering
\caption{\textsc{REAP-GBFS} across the four non-tabletop environments,
each reported on its in-distribution and out-of-distribution settings, as the mean over
three seeds (solve rate $\pm$ standard deviation). For each problem, combined policy and planning time was limited to $300$\,s.}
\label{tab:env:summary}
\small
\begin{tabular}{lccc}
\toprule
Setting & Solve (\%) & Mean exp./steps & MP / plan \\
\midrule
\multicolumn{4}{l}{\emph{Confined-shelf rearrangement}} \\
\quad Monotone (in-distribution)                    & $100.0 \pm 0.0$ & $1.04$ & $8.8$ [7.7]  \\
\quad Non-monotone, $16$ objects (OOD)              & $100.0 \pm 0.0$ & $1.00$ & $11.7$ [12.3] \\
\midrule
\multicolumn{4}{l}{\emph{Multi-level shelf retrieval}} \\
\quad In-distribution ($1$--$5$ blockers)    & $99.3 \pm 1.0$ & $1.20$ & $8.0$ [4.0]  \\
\quad OOD ($4$--$10$ blockers)                      & $100.0 \pm 0.0$ & $1.11$ & $8.7$ [8.0]  \\
\midrule
  \multicolumn{4}{l}{\emph{Access-19}} \\
  \quad Single policy, in-distribution ($\leq 14$ blockers)   & $91.8 \pm 5.8$ & $1.14$ & $37.2$ [8.3]   \\
  \quad Two-phase, $18$-blocker     & $89.8 \pm 2.9$ & $1.36$ & $102.1$ [57.8] \\
\midrule

\multicolumn{4}{l}{\emph{Three-dimensional block stacking}} \\
\quad In-distribution ($1$--$8$ blocks)      & $96.6 \pm 4.8$ & $1.10$ & $14.0$ [3.0]  \\
\quad OOD (combined structures, $4$--$12$ blocks)   & $91.7 \pm 2.4$ & $2.16$ & $85.8$ [28.3] \\
\bottomrule
\end{tabular}
\end{table}

\paragraph{Joint Training.}
  One policy trained jointly on open tabletop,
  confined-shelf and multi-level shelf (two seeds) solves $95.0 \pm 1.3\%$ of the $300$ open-tabletop
  problems, $248$ and $249$ of the $250$
  confined-shelf, and $85$ and $90$ of the $90$ multi-level shelf problems, showing
  that one REAP model can be trained to solve multiple environments.

\subsection{Ablations}
\label{sec:result:ablations}

We swept RL hyperparameters on open tabletop,
with a 40-problem validator and three seeds per value. 
The exploration probability $\varepsilon$ defined in
Section~\ref{sec:method:gnn} is the most impactful: raising it from $0$
to $0.2$ lifts the mean peak validator solve rate from $87.5 \pm 9.4$ to
$96.7 \pm 1.2$ and reduces the spread across seeds. The dead-end cost
penalty improves the solve rate monotonically as it grows from $-22$ to
$-50$. Replay buffer size has no clear effect. We use the
smallest size tried ($20$) because the trainer loop then sees each
collected trace fewer times, biasing training toward recent behavior.
Dead-end cost $-50$ and buffer size $20$ were used in all our experiments.

\subsection{Qualitative Analysis}
\label{sec:result:qualitative}
Across environments, the learned policy shows interpretable regularities
that the domain does not enforce. On hard \emph{open-tabletop}
 instances, it puts blockers into the reachable far corners relative to the target and usually
starts with the southmost occluder, though plans can be sub-optimal with occasional non-blocker shuffles. Trained only on monotone \emph{confined-shelf}
instances, it generalizes to non-monotone configurations, which need parking moves that no
training instance requires.

\section{Limitations}
\label{sec:limitations}
REAP is evaluated entirely in simulation: sim-feasible plans transfer to a
real arm only insofar as the simulated kinematics and geometry are
faithful, and the symbolic state is assumed provided rather than perceived. Each new environment requires its own simulator, a
feasibility oracle supplying per-action verdicts, and a training
problem distribution. Where applicable, online motion-planner queries during training
dominate training cost. The action space
is restricted to prehensile placements on a discrete grid, leaving
non-prehensile manipulation, contact-rich interaction, and
articulated-object handling out of scope.
The clearest failure is Access-19, which appears to require a decomposition into subproblems. A hand-designed split into two phases
solves the task, but a single policy trained end-to-end fails.

\section{Conclusion}
\label{sec:conclusion}
We presented REAP, a framework that learns a relational $Q$-value heuristic
over a coarse symbolic abstraction with a motion planner serving as a
per-action feasibility oracle, and deployed it with greedy best-first
search. The same R-GNN architecture and training loop are used in five
robot manipulation domains.
Geometric feasibility need not be modeled in the symbolic representation to be exploited at search
time. A relational value heuristic can carry it implicitly, and in our experiments
this suffices for the generalization we report.

Future directions include
replacing the motion planner with a learned low-level controller, a
vision-language-action policy, or an RL policy whose success feedback
fills the same per-action oracle role. We also want to incorporate more types
of manipulation (e.g., non-prehensile, loco-manipulation), and the explicit ability for the policy
to identify and use subgoaling. Closing the
sim-to-real gap with physical-robot deployment is an additional research direction.
Finally, non-grid symbolic abstractions are relevant for more complex manipulation tasks, and we want to explore how to learn relational value heuristics over them.

\clearpage
\acknowledgments{The research has been supported by the Alexander von Humboldt Foundation with funds from the German Federal Ministry for Education and Research.
This project was also funded by the German Federal Ministry of Research, Technology and Space (BMFTR) and the Ministry of Culture and Science of the German State of North Rhine-Westphalia (MKW) under the Excellence Strategy of the Federal Government and the L\"ander. This project has received funding from the European Research Council (ERC) under the European Union's Horizon 2020 research and innovation programme (Grant agreement No.\ 885107). Computations were performed with computing resources granted by RWTH Aachen University under project rwth2080.}

\bibliography{paper} %

\clearpage
\appendix

\section{Training Details}
\label{sec:appendix:training-details}

This appendix describes the input graph of the R-GNN, its hyperparameters, and details of
the training loop of Section~\ref{sec:method:gnn}: the training search, the sampling of
training problems, the adaptive search budget, and the choice of checkpoints.
The feasibility checks are described in Appendix~\ref{sec:appendix:envs:shared}.

\begin{figure}[!ht]
\centering
\includegraphics[width=\textwidth]{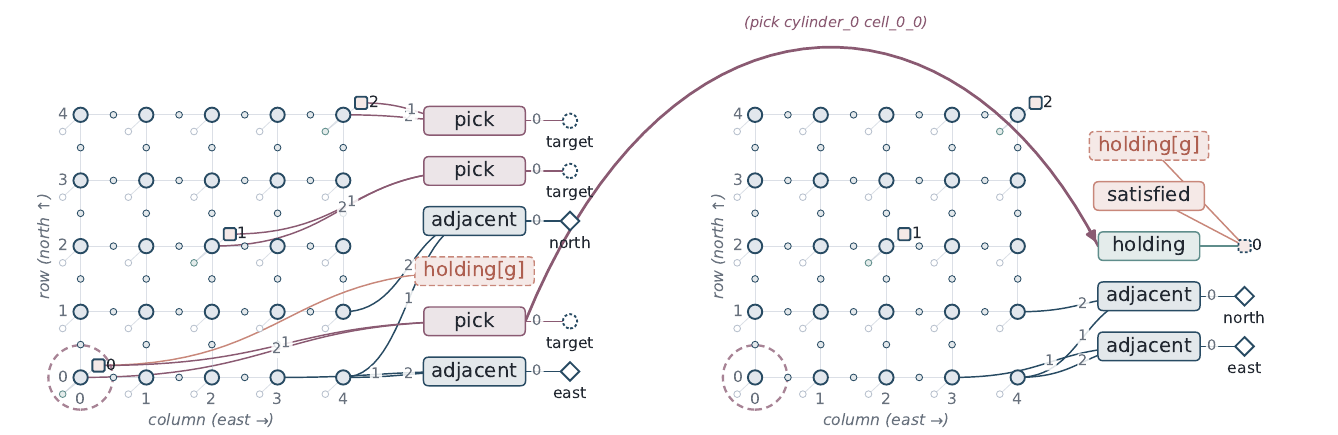}
\caption{Input graph of the R-GNN on a $5\times5$ open-tabletop problem, before (left) and
after (right) the action \texttt{pick(cylinder\_0, cell\_0\_0)}. Circles are cells, squares
are cylinders, rounded boxes are atoms, and edges connect each atom to its arguments,
labeled by argument position. Each applicable action enters the graph as an atom, so a single
pass scores all actions. The graph contains only occupancy, adjacency and the gripper state;
nothing in it says whether an action is geometrically feasible.}
\label{fig:appendix:gnngraph}
\end{figure}

\paragraph{R-GNN Hyperparameters.} In every environment the R-GNN has $L = 30$
message-passing layers with shared weights, Mish activations~\citep{misra2019mish} and
LogSumExp aggregation ($\beta = 12$). The embedding width is $32$ on open tabletop and for
joint training, $64$ on the confined shelf, the multi-level shelf and 3D stacking, and $96$
on Access-19. Table~\ref{tab:appendix:hparams} lists the training hyperparameters.

\begin{table}[h]
\centering
\caption{Training hyperparameters. All runs use the dead-end target $-50$ and a replay buffer
of $20$ chunks. \emph{Budget}: initial value, growth, cap, and the number of consecutive
failures at which growth starts. \emph{Batch}: nodes per network call in the first training
search. \emph{Retries}: retry limit in training (validation); training closes a state at its $N$-th
infeasible action (the first for $N{=}0$), validation allows $N$ retries. \emph{Aux.\ CE}: weight of an
auxiliary cross-entropy loss on the actions of solved plans.}
\label{tab:appendix:hparams}
\small
\resizebox{\linewidth}{!}{%
\begin{tabular}{lccccccc}
\toprule
Environment & Width & $\varepsilon$ & Budget & Time limit & Batch & Retries & Aux.\ CE \\
\midrule
Open tabletop (Table~\ref{tab:opentabletop})$^{a}$ & 32 & 0.2 & 4096, $+1024$, 16384, 5 & 60\,s$^{a}$ & 64 & 2 (2) & -- \\
Sample efficiency & 32 & 0.2 & 4096, $+1024$, 16384, 5 & none & 64 & 0 (1) & -- \\
Confined shelf & 64 & 0.2 & 4096, $+1024$, 16384, 5 & 60\,s & 64 & 2 (2) & 0.1 \\
Joint training & 32 & 0.2 & 4096, $+1024$, 16384, 5 & 60\,s & 64 & 2 (2) & 0.1 \\
Multi-level shelf & 64 & 0 & 64, $\times2$, 2048, 5 & none & 64 & 2 (2) & 0.1 \\
3D stacking & 64 & 0.1 & 64, $\times2$, 4096, 5 & none & 64 & 2 (2) & 0.1 \\
Access-19, single policy & 96 & 0 & 64, $\times2$, 2048, 2 & none & 64 & 2 (2) & 0.1 \\
Access-19, extraction & 96 & 0 & 64, $\times2$, 2048, 2 & none & 4 & 2 (2) & 0.1 \\
Access-19, reload & 96 & 0 & 64, $\times2$, 2048, 2 & none & 1 & 8 (2) & 0.1 \\
\bottomrule
\end{tabular}}

{\small\raggedright $^{a}$ Two of the five Table~\ref{tab:opentabletop} seeds ran without the worker
time limit.\par}
\end{table}

\paragraph{Training Search.} Each training search is first run with batched expansions,
with the number of nodes per network call given in Table~\ref{tab:appendix:hparams}. If it reaches the goal,
it is repeated with one expansion per call, and this second search provides the training
targets. If it fails, only the actions of the start state provide targets. In a state, the
infeasible actions are skipped until the retry limit of Table~\ref{tab:appendix:hparams} is
reached; the action that reaches it receives the dead-end target and closes the state. In the
open-tabletop runs of Table~\ref{tab:opentabletop} and the sample-efficiency runs, failed
searches provide no targets.

\paragraph{Problem Sampling.} Each worker samples training problems with a weight that
depends on the outcome of its last search: \emph{unsolved} (no plan found), \emph{frontier}
(solved with backtracking), or \emph{solved} (solved without backtracking). A frontier problem receives the weight $\omega_m$ for its
first $\tau$ samples, after which the weight decays geometrically towards a floor,
$w = \max\!\big(\omega_{\min},\ \omega_m\,\delta^{\max(0,\,t-\tau)}\big)$, where $t$ counts the
samples since the status last changed. An unsolved problem receives a weight that grows with
its number $\phi$ of consecutive failures, $w = \omega_u\min(1 + \alpha\phi,\ M)$, so that
problems just out of reach keep being tried, and a solved problem keeps a small weight
$w = \omega_s$. We use $\omega_m{=}1$, $\tau{=}20$, $\delta{=}0.9$, $\omega_{\min}{=}0.1$,
$\omega_u{=}0.1$, $\alpha{=}0.1$, $M{=}3$ and $\omega_s{=}0.05$. The open-tabletop runs of
Table~\ref{tab:opentabletop} and the sample-efficiency runs used an earlier, simpler scheme:
weight $1.0$ for the least-sampled frontier problems, $0.1$ for the other frontier problems,
and $0.01$ for solved and unsolved problems.

\paragraph{Adaptive Search Budget.} Each training problem has its own search budget $B$.
$B$ grows, by a fixed increment or by doubling and up to a cap, on every failure that brings
the number of consecutive failures of the problem to a threshold or beyond. A success that
expands only states on the plan halves $B$, but not below its initial value, and any other
success resets $B$ to its initial value. Table~\ref{tab:appendix:hparams} gives these
schedules. The budget concentrates training on the hardest problems that can still be
solved, and we found it crucial for learning long plans. A time limit per search can in
addition stop searches that spend long in motion planning.

\paragraph{Replay Buffer and Checkpoints.} The replay buffer holds the $20$ most recent
chunks of training data. On open tabletop, a checkpoint is saved whenever the result on a
$40$-problem validation set, solved by greedy search, improves, and the last one saved is
evaluated. Joint training uses the final weights; the other environments are described in
Appendix~\ref{sec:appendix:envs}.

\section{Environments}
\label{sec:appendix:envs}

This appendix describes the five environments. We first describe what they share: the
robot, the simulator and the feasibility check. Each environment is then described in
the same order: the scene, the task, the symbolic model, the hidden feasibility
constraints, and the data and evaluation protocol.

\subsection{Shared Backend}
\label{sec:appendix:envs:shared}

\paragraph{Robot and Feasibility Check.} All environments use a single Franka Emika Panda
arm simulated in MuJoCo~\citep{todorov2012mujoco}. The feasibility oracle $X(s) \subseteq
A(s)$ of Section~\ref{sec:method:setup} is a single function \texttt{filter\_action(s, a)}
that tells whether the arm can execute the pick or put motion of action $a$ in state $s$.
Its verdicts are cached for the search on one problem, and the cache is cleared between
problems. The function is implemented in three ways: online motion planning on open
tabletop, online motion planning with exact geometry for 3D stacking, and precomputed
motion libraries for the three shelf environments.

\paragraph{Open Tabletop.} Inverse kinematics is solved with MINK~\citep{zakka2024mink}
(up to three seeds per call) and motions are planned with
RRT-Connect~\citep{kuffner2000rrtconnect}. Collisions are checked for a simplified model of
the hand (capsules for the hand and the last link, and the two finger pads) against the
cylinders and the table, where contacts closer than $1$\,mm count as collisions. The held
cylinder is checked only at the start and end of a put motion; the motion that carries a
picked cylinder away checks neither the cylinder nor the last link. An action is infeasible
when no IK solution or no collision-free path is found within the planning budget. The
planner is not seeded: repeating a check changes the verdict in about $1.2\%$ of the cases,
so the query counts of a problem are single draws.

\paragraph{3D Stacking.} Inverse kinematics is solved analytically over all solution
branches~\citep{he2021franka} along the vertical approach line, and the resulting motions are
checked against the exact meshes of the arm, hand, fingers, blocks and tables with a $2$\,mm
clearance. Motions between grasps are straight joint-space motions when these are
collision-free, and are otherwise planned with the RRT-Connect implementation of
VAMP~\citep{thomason2024vamp} and checked again in MuJoCo. The held block is checked along
the whole motion and may touch its neighbors only while it moves along its own vertical
placement line.

\paragraph{Shelf Environments.} In the confined shelf, the multi-level shelf and Access-19,
every pick and put motion is computed once, offline, and stored together with the set of
cells it collides with. The motions are found with the same analytic IK, checked against the
exact meshes of the arm, the held object and the shelf, including self-collisions of the arm,
and connected to a home configuration with VAMP. An action on a cell is then feasible if some
stored motion for that cell collides with no other occupied cell, so a check is a table
lookup. The three libraries hold $141$, $276$ and $293$ motions and take $15$, $49$ and $1.6$
minutes of wall-clock time to build. On all $240$, $240$ and $700$ random states we tested,
they agree with running the same motion search per query, which validates the precomputation
but not the motion search itself. The libraries are conservative: motions that are not in
the library, such as motions weaving between objects, are never considered.

\paragraph{Clearance.} No motion of an evaluated plan in the shelf environments penetrates an
object, with a minimum clearance of $1.8$\,mm against an intended margin of $2$\,mm, and every solved 3D-stacking plan is verified
by executing it in simulation with an arm clearance of at least $2$\,mm. The open-tabletop
check uses the simplified hand model above and is not covered by these statements.

\paragraph{Motion-Planner Queries.} \emph{MP/plan} counts the distinct $(s,a)$ queries per
solved problem: planner calls on open tabletop and in 3D stacking, and library lookups in the
shelf environments.

\subsection{Open-Tabletop Rearrangement}
\label{sec:appendix:envs:tabletop}

\begin{figure}[!t]
\centering
\includegraphics[width=0.85\linewidth]{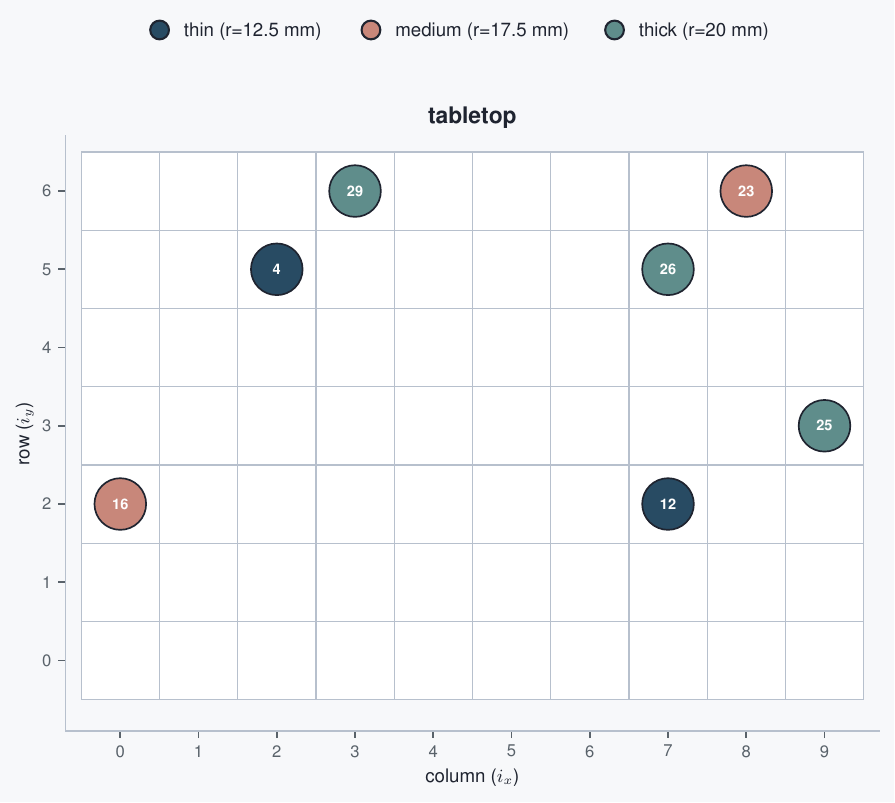}
\\[2mm]
\small\begin{tabular}{@{}ll@{}}
\toprule
Workspace footprint & $40\times28$\,cm table \\
\midrule
\multicolumn{2}{@{}l}{\emph{Grid regions} (resolution over extent, cell pitch)}\\
\multicolumn{2}{@{}l}{\quad table: $10\times7$ over $40\times28$\,cm, pitch $40$\,mm}\\
\midrule
\multicolumn{2}{@{}l}{\emph{Object classes}}\\
\multicolumn{2}{@{}l}{\quad thin cyl ($\times$15): $r=12.5$\,mm, $h=8$\,cm, 1 cell}\\
\multicolumn{2}{@{}l}{\quad medium cyl ($\times$10): $r=17.5$\,mm, $h=10$\,cm, 1 cell}\\
\multicolumn{2}{@{}l}{\quad thick cyl ($\times$5): $r=20$\,mm, $h=12$\,cm, 1 cell}\\
\bottomrule
\end{tabular}

\caption{Open tabletop: the $10\times7$ grid with an example state, and the dimensions of
the workspace and objects.}
\label{fig:appendix:env:tabletop}
\end{figure}

\paragraph{Scene.}
A $40\times28$\,cm table in front of the arm is divided into a $10\times7$ grid of $40$\,mm
cells (Figure~\ref{fig:appendix:env:tabletop}). Cylinders of three sizes stand on the
table, each in one cell. The hidden constraint is that neighboring cylinders can
block the gripper as it approaches or retracts.

\paragraph{Task.}
The goal is a single atom \texttt{holding(o)}: grasp a target cylinder. The $600$ training
problems have $3$ to $14$ cylinders: $480$ with random placements, $60$ in which the target
can only be reached after relocating a chain of $K{=}2$ blockers, and $60$ with $K{=}3$. The
$300$ evaluation problems have $10$ to $24$ cylinders, reference plans of up to $15$ actions,
and chains with $K{=}2$ and $K{=}3$.

\begin{figure}[!ht]
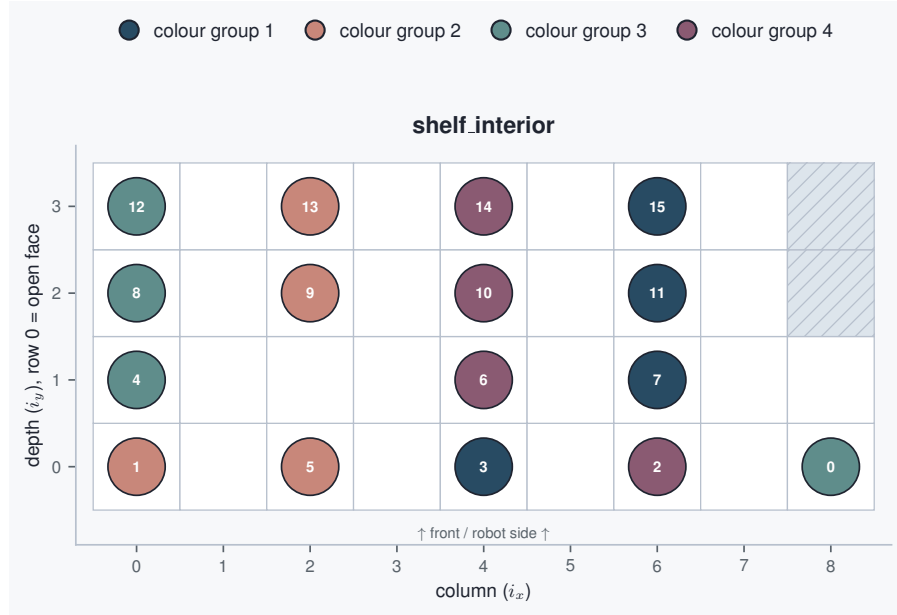

\centering
\begin{minipage}[t]{0.56\linewidth}
\scriptsize
\begin{verbatim}
(define (domain tabletop-manipulation)
  (:requirements :strips :typing
    :negative-preconditions :derived-predicates
    :disjunctive-preconditions
    :existential-preconditions)
  (:types cell cylinder direction)
  (:constants north east - direction)
  (:predicates
    (adjacent ?dir - direction ?cel1 - cell
              ?cel2 - cell)
    (occupied ?cel - cell ?cyl - cylinder)
    (empty ?cel - cell)
    (holding ?cyl - cylinder)
    (gripper-empty)
    (discarded ?cyl - cylinder))
  (:action pick
    :parameters (?cyl - cylinder ?cel - cell)
    :precondition (and (gripper-empty)
                       (occupied ?cel ?cyl))
    :effect (and (not (gripper-empty))
                 (not (occupied ?cel ?cyl))
                 (empty ?cel) (holding ?cyl)))
  (:action put
    :parameters (?cyl - cylinder ?cel - cell)
    :precondition (and (holding ?cyl)
                       (empty ?cel))
    :effect (and (gripper-empty)
                 (occupied ?cel ?cyl)
                 (not (empty ?cel))
                 (not (holding ?cyl)))))
\end{verbatim}
\end{minipage}\hfill
\begin{minipage}[t]{0.42\linewidth}
\scriptsize
\begin{verbatim}
(define (problem config-1)
  (:domain tabletop-manipulation)
  (:objects
    cell_0_0 cell_0_1 ... cell_9_6
      - cell
    cylinder_0 cylinder_1 ...
      - cylinder)
  (:init
    (adjacent north cell_0_0 cell_0_1)
    (adjacent east cell_0_0 cell_1_0)
    ...
    (occupied cell_5_0 cylinder_0)
    (occupied cell_5_3 cylinder_1)
    ...
    (empty cell_0_0)
    ...
    (gripper-empty))
  (:goal (and
    (holding cylinder_1))))
\end{verbatim}
\end{minipage}
\caption{Open-tabletop PDDL: the domain (left, without comments; \texttt{discarded} is not
used) and an abbreviated problem with $10$ cylinders (right).}
\label{fig:appendix:tabletop:pddl}
\end{figure}

\paragraph{Symbolic Model.}
The predicates are \texttt{occupied(c,cyl)}, \texttt{empty(c)}, \texttt{holding(cyl)},
\texttt{gripper-empty} and the static \texttt{adjacent(dir,c,c')} for the north and east
neighbors of a cell. The action \texttt{pick(cyl,c)} requires an empty gripper and
\texttt{cyl} in \texttt{c}, and \texttt{put(cyl,c)} requires holding \texttt{cyl} and an
empty \texttt{c}. With $N$ cylinders and $C = 70$ cells, a state has $N$ applicable actions
when the gripper is empty and $C - N + 1$ when it holds a cylinder.
Figure~\ref{fig:appendix:tabletop:pddl} shows the domain and an example problem.

\paragraph{Feasibility.}
Picks use a side grasp from the front of the table, and puts approach the cell from above.
An action is infeasible when the motion check of
Section~\ref{sec:appendix:envs:shared} finds no collision-free motion, typically because a
cylinder stands in front of the target cell in the same or an adjacent column.

\paragraph{Data and Evaluation.}
Reference plans of the training problems have up to $13$ actions. Each evaluation problem
has a $300$\,s cap, and a problem counts as solved when the returned plan passes the
feasibility check in every step. Figure~\ref{fig:ood} pools the five seeds per bucket, with
$20$ problems per object count and $240$, $20$ and $40$ problems in the random, $K{=}2$ and
$K{=}3$ buckets.

\subsection{Confined-Shelf Rearrangement}
\label{sec:appendix:envs:shelf}

\begin{figure}[!t]
\centering
\includegraphics[width=0.85\linewidth]{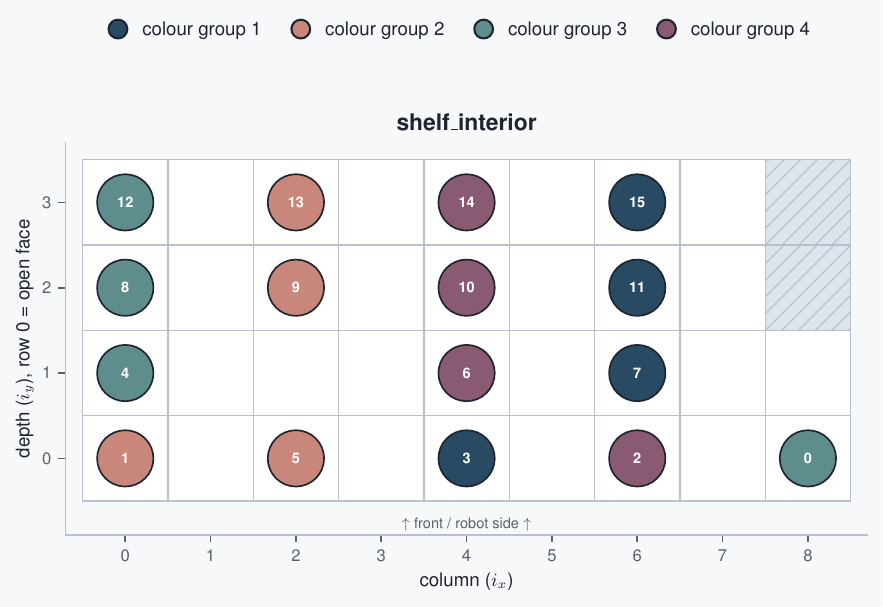}
\\[2mm]
\small\begin{tabular}{@{}ll@{}}
\toprule
Workspace & closed cubicle, inner cavity $79\times44\times34$\,cm, open $-y$ face \\
\midrule
\multicolumn{2}{@{}l}{\emph{Grid regions} (resolution over extent, cell pitch)}\\
\multicolumn{2}{@{}l}{\quad shelf\_interior: $9\times4$ minus 2 deep back-right cells $=34$ cells over $63\times28$\,cm, pitch $70$\,mm}\\
\midrule
\multicolumn{2}{@{}l}{\emph{Object classes}}\\
\multicolumn{2}{@{}l}{\quad cylinder (up to 16, 2--4 color groups): $r=30$\,mm, $h=20$\,cm, 1 cell}\\
\bottomrule
\end{tabular}

\caption{Confined shelf: the $9\times4$ grid, open at $-y$, with an example state, and the
dimensions of the workspace and objects.}
\label{fig:appendix:env:shelf}
\end{figure}

\paragraph{Scene.}
A cubicle, closed on five sides and open at the front, holds up to sixteen identical
bottle-sized cylinders in two to four colors (Figure~\ref{fig:appendix:env:shelf}). Its
floor is divided into a $9\times4$ grid of $70$\,mm cells; the two back-right cells cannot be
reached, which leaves $34$ cells. The hidden constraint is that every grasp approaches from
the open face, past the cylinders in front of the target.

\paragraph{Task.}
The goal is to sort the cylinders by color, with each color group in its own column.
Training problems are \emph{monotone}, with $3$ to $9$ cylinders that can each be moved
directly to a final column. The evaluation uses monotone problems of the same size
(in-distribution) and \emph{non-monotone} problems with $10$, $12$, $14$ and $16$ cylinders
(out-of-distribution), in which at least one cylinder must first be parked in an
intermediate cell.

\paragraph{Symbolic Model.}
The predicates and actions are those of open tabletop, plus a static relation
\texttt{color-of(cyl,color)}. The \texttt{adjacent} and \texttt{color-of} atoms appear in
no precondition; they only give the R-GNN structure to reason with. With $N$ cylinders and
$C = 34$ cells, a state has $N$ or $C - N + 1$ applicable actions.

\paragraph{Feasibility.}
With the motion library, the check reduces to a rule over cells: a pick or put at a cell is
feasible if its two neighbors in the same row and every cell in front of it in the same and
the two adjacent columns are empty, and, for back-row cells in columns $3$ to $7$, the front
cell two columns to the left, which the forearm sweeps, is empty as well. Cells behind the
target never block.

\paragraph{Data and Evaluation.}
The $300$ training problems have reference plans of $4$ to $66$ actions (median $8$), found by
a planner that is aware of the feasibility check but not optimal. Each evaluation set has
$50$ problems; the in-distribution set runs with a $300$\,s cap and the non-monotone sets
with a $100$\,s cap. One $14$-cylinder problem whose goal cannot be reached under the
feasibility check, as exhaustive search shows, was replaced by the next solvable problem
from the generator that is not isomorphic to another one. The
in-distribution set is also the validation set during training, so we report the final
weights of all three seeds. Table~\ref{tab:env:summary} shows the in-distribution set and
the $16$-cylinder set; Table~\ref{tab:appendix:shelf:fullsweep} shows all non-monotone sets.
\textsc{REAP-GBFS} solves $98$--$100\%$ at every size with almost no search overhead.

\begin{table}[h]
\centering
\caption{\textsc{REAP-GBFS} on the non-monotone confined-shelf problems (three seeds, final
weights).}
\label{tab:appendix:shelf:fullsweep}
\small
\begin{tabular}{lccc}
\toprule
Cylinders & Solve (\%) & Mean exp./steps & MP / plan \\
\midrule
$10$ & $98.0 \pm 2.8$ & $1.10$ & $16.8$ [12.7] \\
$12$ & $98.0 \pm 1.6$ & $1.03$ & $12.2$ [12.7] \\
$14$ & $100.0 \pm 0.0$ & $1.01$ & $12.0$ [12.3] \\
$16$ & $100.0 \pm 0.0$ & $1.00$ & $11.7$ [12.3] \\
\bottomrule
\end{tabular}
\end{table}

\subsection{Multi-Level Shelf Retrieval}
\label{sec:appendix:envs:retrieval}

\begin{figure}[!t]
\centering
\includegraphics[width=0.85\linewidth]{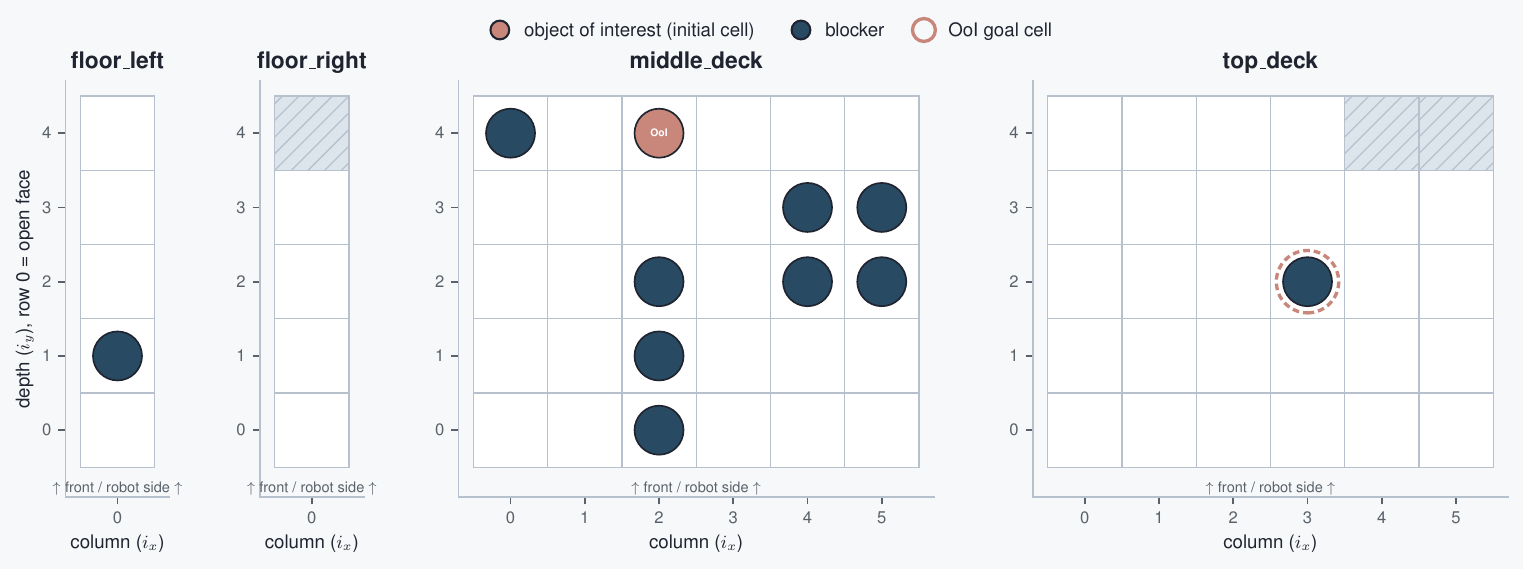}
\\[2mm]
\small\begin{tabular}{@{}p{0.95\linewidth}@{}}
\toprule
\emph{Workspace}: open two-deck shelf (no walls), decks $80\times45$\,cm centred at $(0.40, 0.58)$\,m, surfaces $z=0.34/0.58$\,m; floor split by a separator; robot base at the origin, front edge $y=0.355$\,m \\
\midrule
\emph{Grid regions} (columns $\times$ rows over extent, cell pitch)\\
\quad floor\_left: $1\times5$ = 5 cells over $7.0\times35.0$\,cm, pitch $70$\,mm\\
\quad floor\_right: $1\times5$ minus 0\_4 = 4 cells over $7.0\times35.0$\,cm, pitch $70$\,mm\\
\quad middle\_deck: $6\times5$ = 30 cells over $42.0\times35.0$\,cm, pitch $70$\,mm\\
\quad top\_deck: $6\times5$ minus 4\_4, 5\_4 = 28 cells over $42.0\times35.0$\,cm, pitch $70$\,mm\\
\midrule
\emph{Objects}\\
\quad YCB-proxy box $6\times6\times9$\,cm, 1 cell; grasp: horizontal front grasp $2.5$\,cm above the box centre\\
\quad sample (shown): 10 blockers + 1 OoI, 67 cells in total\\
\bottomrule
\end{tabular}

\caption{Multi-level shelf: the four grid regions with an example state, and the dimensions
of the workspace and objects.}
\label{fig:appendix:env:retrieval}
\end{figure}

\paragraph{Scene.}
An open shelf with two $80\times45$\,cm decks stands in front of the arm, with a floor
compartment below that is split in two (Figure~\ref{fig:appendix:env:retrieval}). It is
divided into four grids of $70$\,mm cells: the middle deck ($30$ cells), the top deck
($28$), and the left and right floor segments ($5$ and $4$), $C = 67$ cells in total. Each
problem has one object of interest (OoI) and several box-shaped blockers, each in one cell.
All objects are grasped from the front with a horizontal hand. The hidden constraint is that
neighboring objects can block this grasp, either along the approach or where the forearm
sweeps.

\paragraph{Task.}
The goal is to move the OoI to a target cell elsewhere in the shelf, which requires
relocating the blockers that obstruct its grasp or its target. Training problems have $1$
to $5$ blockers (mean $2.9$), and the out-of-distribution problems have $4$ to $10$ (mean
$6.8$).

\paragraph{Symbolic Model.}
The predicates and actions are those of open tabletop over the four regions, with
\texttt{adjacent} only within a region; \texttt{adjacent} appears in no precondition. With
$M$ objects, a state has $M$ or $C - M + 1$ applicable actions.

\paragraph{Feasibility.}
With the motion library, a floor cell is blocked only by objects in front of it in the same
column. A deck cell needs its two neighbors in the same row and every cell in front of it,
within one column, to be empty; cells two or more rows deep also need some cells two columns
away to be empty, which the forearm sweeps. Objects behind the target or in another region
never block.

\paragraph{Data and Evaluation.}
The $276$ training problems have reference plans of $4$ to $10$ actions from a planner that
is aware of the feasibility check. They were taken from an earlier version of the
environment, keeping only the problems this feasibility check can solve. Each problem has a
$300$\,s cap, and we report the final weights of three seeds. The original in-distribution
set is also the validation set during training and shares problems with the training set,
so Table~\ref{tab:env:summary} uses an in-distribution set of $46$ problems with no problem
in common with training ($35$ original problems and $11$ new ones); the original set gives the
same solve rate. The
out-of-distribution set has $44$ problems; two more were removed because exhaustive search
shows that they cannot be solved under the feasibility check.

\subsection{Access-19}
\label{sec:appendix:envs:access19}

\begin{figure}[!t]
\centering
\includegraphics[width=0.85\linewidth]{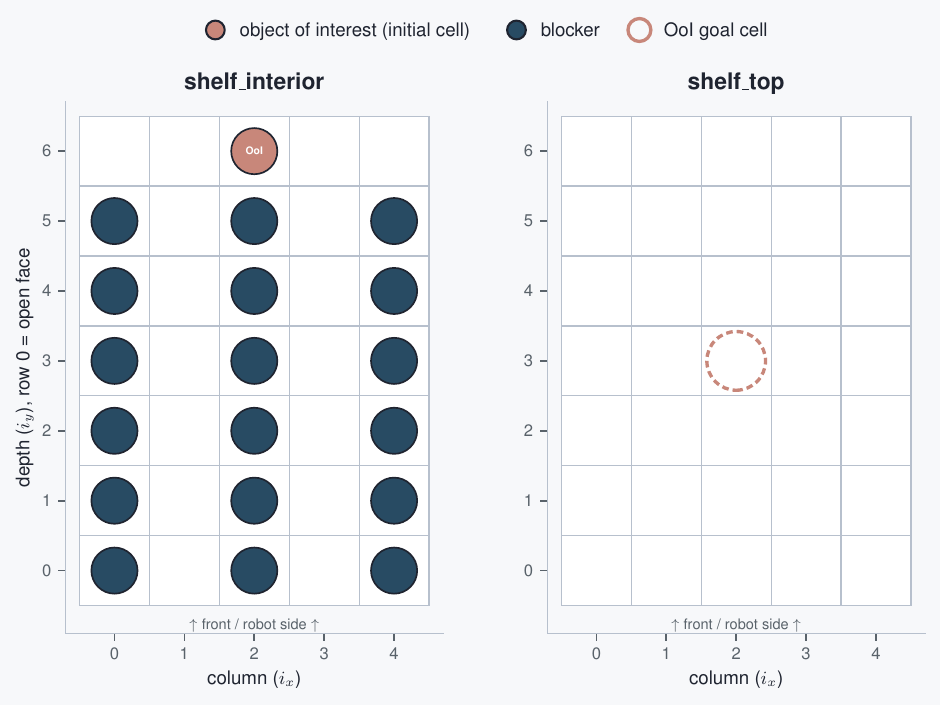}
\\[2mm]
\small\begin{tabular}{@{}p{0.95\linewidth}@{}}
\toprule
\emph{Workspace}: bookcase centred on the robot axis, front edge $y=0.375$\,m; compartment $39\times27.8\times24$\,cm (floor $z=0.34$\,m), open top deck at $z=0.592$\,m \\
\midrule
\emph{Grid regions} (columns $\times$ rows over extent, cell pitch)\\
\quad shelf\_interior: $5\times7$ = 35 cells over $20.0\times25.2$\,cm, pitch $40\times36$\,mm\\
\quad shelf\_top: $5\times7$ = 35 cells over $20.0\times25.2$\,cm, pitch $40\times36$\,mm\\
\midrule
\emph{Objects}\\
\quad box $3\times3\times10$\,cm, 1 cell; one grasp: hand horizontal, $7$\,mm below the box top, approach $+y$\\
\quad sample (shown): 18 blockers + 1 OoI (OoI goal: shelf\_top\_\_2\_3)\\
\bottomrule
\end{tabular}

\caption{Access-19: the two grid regions with the OoI (salmon) among the blockers (dark
blue) and its goal cell (ringed), and the dimensions of the workspace and objects.}
\label{fig:appendix:env:access19}
\end{figure}

\paragraph{Scene.}
Access-19 is a dense bookcase with an open front whose roof serves as a top deck
(Figure~\ref{fig:appendix:env:access19}). Each of the two levels is divided into $5\times7$
cells, $C = 70$ cells in total. The OoI and all blockers are boxes of the same size, each in
one cell, and all are grasped from the front.

\paragraph{Task.}
The OoI starts deep inside the bookcase among $0$ to $18$ blockers, and the goal moves it to
a fixed cell on the top deck. In \emph{extraction} problems this is the whole goal. In
\emph{return} problems, every blocker must also end in its initial cell, so blockers must be
moved away from their goal cells and later put back. The $18$-blocker problem is adapted from
\citet{bouhsain2025longhorizon}; its eight benchmark files are one layout up to the naming of
the blockers. We also generate $72$ new $18$-blocker layouts, by placing the OoI in one of
$11$ deep cells and the blockers in $18$ of the $20$ remaining cells of a fixed pattern, that
differ from all training and evaluation problems and from each other.

\paragraph{Symbolic Model.}
The predicates and actions are those of the multi-level shelf, plus the derived predicate
\texttt{blocking(obj,c)}, true when \texttt{obj} is in front of cell \texttt{c} in the same
column. It is computed from the other atoms, appears in no precondition, and only gives the
R-GNN an occlusion signal.

\paragraph{Feasibility.}
With the motion library, a cell needs its two neighbors in the same row to be empty, since
the hand housing, about $21$\,cm wide, passes through the top of every box it travels past. Cells behind the front row also need every
cell in front of them, within two columns, to be empty, and in the two rightmost columns,
cells two or more rows deep also need the cells in front of them three columns to the left
to be empty. Cells behind the target and on the other level never block.

\paragraph{Single Policy.}
The single goal-conditioned policy is trained on $358$ problems with $0$ to $14$ blockers,
$207$ of them return problems, with reference plans of $2$ to $104$ actions (median $14$).
Its checkpoint is selected on a separate set of $93$ problems that shares no layout with any
training or evaluation set.

\paragraph{Two-Phase Pipeline.}
The pipeline splits a problem into extraction and reload and trains one policy for each, with
the same architecture and training loop as before (hyperparameters in
Table~\ref{tab:appendix:hparams}). The \emph{extraction policy} is trained on $609$ extraction
problems with $0$ to $18$ blockers, including the $18$-blocker benchmark layout; $315$ of them
start from a state reached part-way along a long reference plan, $8$, $16$ or $24$ actions
before its end. The \emph{reload policy} is trained on $706$ problems with $2$ to $17$
blockers: $215$ start from the state after extraction, and $491$ start $8$, $16$ or $24$
actions before the end of a reload plan. Its training search expands one node at a time,
since batched expansion limits the search depth to the budget divided by the batch size,
which is shorter than reload plans of $30$ or more actions. Both policies are trained with
three seeds and paired by seed; the extraction policy is used with its final weights and the
reload policy with its best checkpoint on a validation set of $34$ problems that shares no
layout with any evaluation set. At test time, the extraction policy runs with the OoI cell
as its goal, the reload policy continues from the reached state towards the full goal, both
within one time cap, and the combined plan is checked from the initial state.

\paragraph{Evaluation.}
The in-distribution set has $93$ problems with $0$ to $14$ blockers, $22$ of them return
problems. The out-of-distribution set has $30$ problems, $22$ with up to $14$ blockers and
the eight $18$-blocker benchmark files. Both use a $100$\,s cap, and the $72$ new
$18$-blocker layouts use a $300$\,s cap. On the $81$ in-distribution problems that neither
two-phase policy saw in training, the pipeline solves $95.5 \pm 3.8\%$.

\subsection{Three-Dimensional Block Stacking}
\label{sec:appendix:envs:stacking}

\begin{figure}[!t]
\centering
\includegraphics[width=0.85\linewidth]{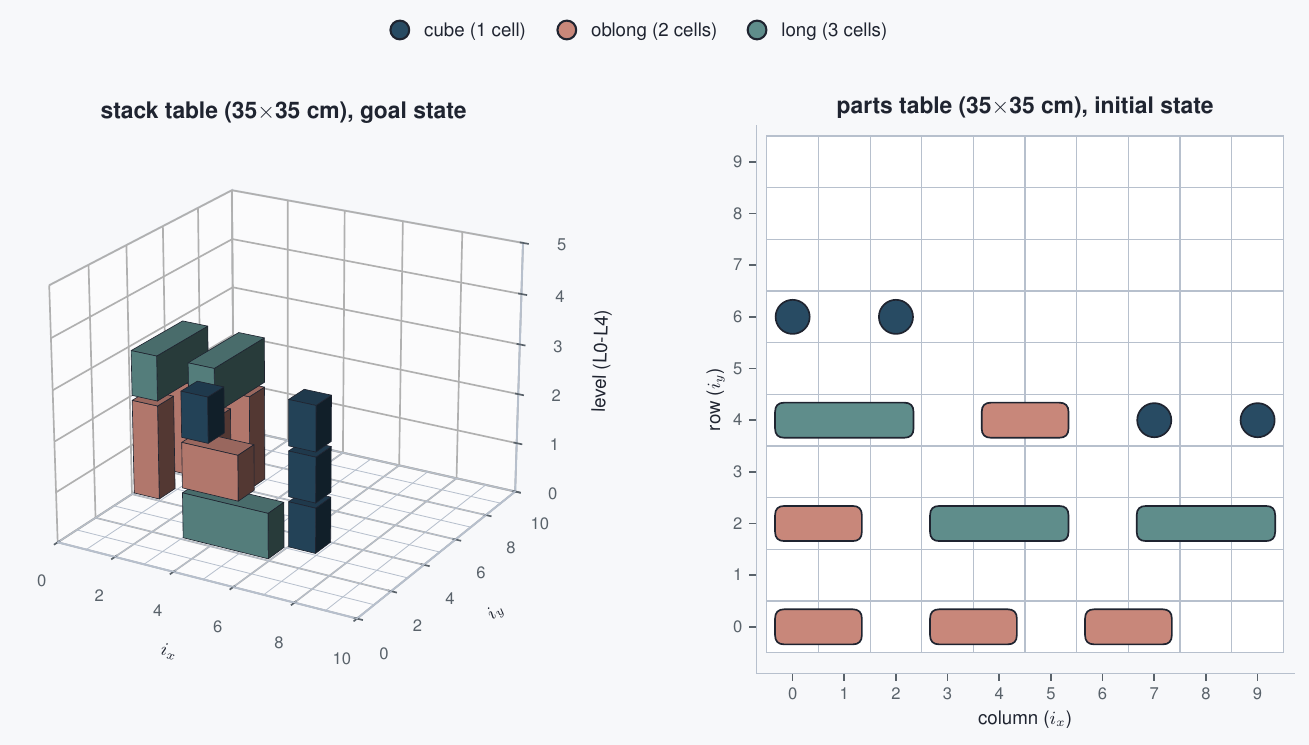}
\\[2mm]
\small\begin{tabular}{@{}ll@{}}
\toprule
Workspace footprint & two $35\times35$\,cm tables; stack $10\times10\times5$ grid, parts $10\times10$ grid \\
\midrule
\multicolumn{2}{@{}l}{\emph{Grid regions} (resolution over extent, cell pitch)}\\
\multicolumn{2}{@{}l}{\quad stack (per level, $\times$5): $10\times10$ over $30\times30$\,cm, pitch $30$\,mm}\\
\multicolumn{2}{@{}l}{\quad parts: $10\times10$ over $30\times30$\,cm, pitch $30$\,mm}\\
\midrule
\multicolumn{2}{@{}l}{\emph{Object classes} (pool of 38 per problem)}\\
\multicolumn{2}{@{}l}{\quad cube ($\times$20): $30$\,mm edge, 1 cell}\\
\multicolumn{2}{@{}l}{\quad oblong ($\times$12): $60\times30\times30$\,mm, 2 cells (flat or upright)}\\
\multicolumn{2}{@{}l}{\quad long ($\times$6): $90\times30\times30$\,mm, 3 cells (flat or upright)}\\
\bottomrule
\end{tabular}

\caption{3D stacking: a $12$-block goal structure on the stack table (left), the blocks on
the parts table in the initial state (right), and the dimensions of the workspace and blocks.}
\label{fig:appendix:env:stacking}
\end{figure}

\paragraph{Scene.}
Two $35\times35$\,cm tables stand in front of and behind the arm
(Figure~\ref{fig:appendix:env:stacking}). On the \emph{parts table}, blocks lie flat on a
$10\times10$ grid. The \emph{stack table} has a $10\times10\times5$ grid, five layers of
cells stacked one block height apart. Cells are $30$\,mm wide, the edge length of a cube, so
neighboring blocks touch.

\paragraph{Task.}
There are three block shapes: cubes (one cell), oblongs (two cells) and longs (three cells).
Each problem declares a pool of $38$ blocks ($20$ cubes, $12$ oblongs and $6$ longs), of which
$1$ to $8$ take part in a training problem; the others are hidden in the simulator. The goal
gives the target position and orientation of each participating block on the stack table.
The out-of-distribution problems combine two or more simpler structures into one structure
of $4$ to $12$ blocks.

\paragraph{Symbolic Model.}
Oblongs and longs can lie flat along $x$, lie flat along $y$, or stand upright on the stack
table. Cubes and flat blocks are picked from above; upright blocks are picked from the front,
near one end. Besides pick and put actions for each shape, the domain has actions that change
the orientation of the held block in the air. A further atom records whether the held block
is gripped at its center or at an end; orientation changes keep it, every put clears it, and
no precondition reads it. The preconditions ensure support: a block can be picked only if the cell above it is
empty, and placed only on the table or directly on top of another block.

\paragraph{Feasibility.}
Whether the gripper and the arm collide with neighboring blocks is not modeled in PDDL; it is
decided by the online check of Section~\ref{sec:appendix:envs:shared}.

\paragraph{Data and Evaluation.}
The $630$ training problems have $1$ to $8$ target blocks in five difficulty levels ($140$,
$160$, $120$, $120$ and $90$ problems), with reference plans of $2$ to $22$ actions,
and a separate set of $99$ problems is used for checkpoint selection. $619$ of the training
problems come from an earlier version of the problem set; the other $11$ replace problems
that cannot be solved under the current feasibility check. The in-distribution set has $95$
problems over the five levels. Six of its hardest problems also appear in the
checkpoint-selection set, so Table~\ref{tab:env:summary} reports the other $89$
($94.7 \pm 7.4\%$ on all $95$). The out-of-distribution set has $20$ problem files
($19$ distinct problems) with $4$ to $12$ blocks. Each problem has a $300$\,s cap. One seed never learns the hardest level
($6$ to $8$ blocks), which accounts for all its in-distribution failures. With the final
weights instead of the selected checkpoints (the last save with the best validation result), the solve rates are $91.0 \pm 12.7\%$
(in-distribution) and $90.0 \pm 4.1\%$ (out-of-distribution).

\section{Open-Tabletop Ablations and Sample Efficiency}
\label{sec:appendix:ablations}

All studies in this appendix train on the open-tabletop training set. For training runs we
report the solve rate on the $40$-problem validation set, both its peak during training
(\emph{max}), which a run would report when keeping its best checkpoint, and its value in the
last validation pass (\emph{last}). The \emph{combined configuration}, used for
Table~\ref{tab:opentabletop} and the sample-efficiency study, is $\varepsilon{=}0.2$, dead-end
target $-50$ and a buffer of $20$ chunks, with or without a $60$\,s time limit per training
search (Table~\ref{tab:appendix:hparams}).

\paragraph{Single-Axis Ablations.} Table~\ref{tab:appendix:ablations} varies one
hyperparameter at a time: the exploration probability $\varepsilon$, the dead-end target, the
time limit per training search, the cap of the search budget, and the buffer size. Each
setting is trained with three seeds for $24$ hours. Most runs were stopped by this limit
rather than converging, and the gap between \emph{max} and \emph{last} suggests that late
training overfits.

\begin{table}[h]
\centering
\caption{Single-axis ablations: peak validation solve rate (\%) of each seed, mean of the
peak (\emph{max}) and last (\emph{last}) solve rates, and the number of training
iterations.}
\label{tab:appendix:ablations}
\small
\begin{tabular}{lcccccc}
\toprule
Setting & Seed 1 & Seed 2 & Seed 3 & Mean max & Mean last & Iters \\
\midrule
\multicolumn{7}{l}{\emph{Exploration probability $\varepsilon$}} \\
$0.0$  & 75.0 & 97.5 & 90.0 & $87.5 \pm 9.4$ & $73.3 \pm 7.7$ & 1304 \\
$0.05$ & 97.5 & 92.5 & 92.5 & $94.2 \pm 2.4$ & $88.3 \pm 6.2$ & 1159 \\
$0.1$  & 95.0 & 92.5 & 95.0 & $94.2 \pm 1.2$ & $85.8 \pm 5.1$ & 1356 \\
$0.2$  & 97.5 & 97.5 & 95.0 & $96.7 \pm 1.2$ & $93.3 \pm 2.4$ & 1389 \\
\midrule
\multicolumn{7}{l}{\emph{Dead-end target}} \\
$-22$ & 80.0 & 82.5 & 77.5 & $80.0 \pm 2.0$ & $65.8 \pm 3.1$ & 1710 \\
$-30$ & 75.0 & 80.0 & 92.5 & $82.5 \pm 7.4$ & $72.5 \pm 8.2$ & 1719 \\
$-50$ & 92.5 & 87.5 & 90.0 & $90.0 \pm 2.0$ & $76.7 \pm 7.2$ & 1689 \\
\midrule
\multicolumn{7}{l}{\emph{Time limit per search (s; $0$: none)}} \\
$0$   & 72.5 & 75.0 & 75.0 & $74.2 \pm 1.2$ & $64.2 \pm 1.2$ & 853 \\
$60$  & 77.5 & 97.5 & 95.0 & $90.0 \pm 8.9$ & $75.8 \pm 8.3$ & 1731 \\
$300$ & 75.0 & 72.5 & 77.5 & $75.0 \pm 2.0$ & $64.2 \pm 1.2$ & 1250 \\
\midrule
\multicolumn{7}{l}{\emph{Budget cap (multiple of the initial budget)}} \\
$2$ & 97.5 & 80.0 & 77.5 & $85.0 \pm 8.9$ & $74.2 \pm 7.7$ & 1749 \\
$4$ & 75.0 & 77.5 & 92.5 & $81.7 \pm 7.7$ & $66.7 \pm 11.6$ & 1732 \\
$8$ & 92.5 & 95.0 & 95.0 & $94.2 \pm 1.2$ & $79.2 \pm 10.1$ & 1309 \\
\midrule
\multicolumn{7}{l}{\emph{Buffer size (chunks)}} \\
$20$  & 77.5 & 90.0 & 77.5 & $81.7 \pm 5.9$ & $72.5 \pm 3.5$ & 3734 \\
$57$  & 72.5 & 77.5 & 85.0 & $78.3 \pm 5.1$ & $70.0 \pm 2.0$ & 1224 \\
$100$ & 75.0 & 77.5 & 85.0 & $79.2 \pm 4.2$ & $75.0 \pm 4.1$ & 805 \\
\bottomrule
\end{tabular}
\end{table}

\paragraph{Combined Configuration.} Table~\ref{tab:appendix:combo} shows an earlier set of
runs with the combined configuration, three seeds with and three without the time limit,
measured on the validation set after about $11$ hours of training.
Table~\ref{tab:appendix:combo:eval} evaluates the runs of Table~\ref{tab:opentabletop} on the
$300$ test problems, per seed, from checkpoints saved $23$ to $27$ hours into training. One
seed without the time limit is excluded there because it overtrained (early stopping did not
fire in time).

\begin{table}[h]
\centering
\caption{Earlier runs of the combined configuration: validation solve rate (\%) after about
$11$ hours.}
\label{tab:appendix:combo}
\small
\begin{tabular}{lccc}
\toprule
Run & Peak (max) & Final (last) & Iters \\
\midrule
\multicolumn{4}{l}{\emph{No time limit}} \\
Seed 1 & 95.0 & 87.5 & 1372 \\
Seed 2 & 95.0 & 90.0 & 2005 \\
Seed 3 & 95.0 & 82.5 & 1920 \\
Mean   & $95.0 \pm 0.0$ & $86.7 \pm 3.1$ & 1766 \\
\midrule
\multicolumn{4}{l}{\emph{Time limit $60$\,s}} \\
Seed 1 & 90.0 & 87.5 & 2526 \\
Seed 2 & 97.5 & 77.5 & 2530 \\
Seed 3 & 92.5 & 82.5 & 1785 \\
Mean   & $93.3 \pm 3.1$ & $82.5 \pm 4.1$ & 2280 \\
\midrule
All $6$ seeds & $94.2 \pm 2.4$ & $84.6 \pm 4.4$ & 2023 \\
\bottomrule
\end{tabular}
\end{table}

\begin{table}[h]
\centering
\caption{The five runs of Table~\ref{tab:opentabletop} on the $300$ test problems. Columns
as in Table~\ref{tab:opentabletop}; \emph{Plan len.} is the mean length of the solved plans.}
\label{tab:appendix:combo:eval}
\small
\begin{tabular}{lccccc}
\toprule
 & \multicolumn{3}{c}{\textsc{REAP-GBFS}} & \multicolumn{2}{c}{\textsc{REAP-greedy}} \\
\cmidrule(lr){2-4}\cmidrule(lr){5-6}
Run & Solve (\%) & Exp./steps & MP/plan & Solve (\%) & Plan len. \\
\midrule
\multicolumn{6}{l}{\emph{No time limit}} \\
Seed 2 & 96.0 & 1.19 & 13.1 [3] & 90.7 & 4.6 \\
Seed 3 & 96.7 & 1.18 & 11.5 [3] & 89.3 & 4.2 \\
Mean   & $96.3$ & 1.19 & 12.3 & $90.0$ & 4.4 \\
\midrule
\multicolumn{6}{l}{\emph{Time limit $60$\,s}} \\
Seed 1 & 95.0 & 1.36 & 19.8 [3] & 86.7 & 6.3 \\
Seed 2 & 96.0 & 1.27 & 22.8 [3] & 91.7 & 4.6 \\
Seed 3 & 96.0 & 1.45 & 18.1 [3.5] & 89.3 & 5.0 \\
Mean   & $95.7 \pm 0.5$ & 1.36 & 20.2 & $89.2 \pm 2.0$ & 5.3 \\
\bottomrule
\end{tabular}
\end{table}

\paragraph{Sample Efficiency.}
\label{sec:appendix:samplesize}
To see how much training data REAP needs, we train the combined configuration (without time
limit and with training retry limit $0$, Table~\ref{tab:appendix:hparams}) on nested subsets
of the training set with $N = 100$, $200$ and $400$ problems, three seeds each, and evaluate
each run with \textsc{REAP-GBFS} on the $300$ test problems
(Table~\ref{tab:appendix:samplesize}). The solve rate hardly depends on $N$: $87.7\%$ at
$N{=}100$ and $90.8\%$ at $N{=}400$, with the lower value at $N{=}200$ within the spread
across seeds. More data mainly makes the search more efficient: the search overhead falls
from $2.44$ to $2.17$ and $1.99$ expansions per plan step. The results in
Table~\ref{tab:opentabletop} use all $600$ problems.

\begin{table}[h]
\centering
\caption{Sample efficiency: \textsc{REAP-GBFS} trained on $N$ problems (three seeds each).
Columns as in Table~\ref{tab:opentabletop}.}
\label{tab:appendix:samplesize}
\small
\begin{tabular}{lccccc}
\toprule
$N$ & Solve (\%) & MP / plan & Mean exp./steps & Median exp./steps & Time (s) \\
\midrule
$100$ & $87.7 \pm 2.4$ & 49.8 [5.2] & 2.44 & 1.18 & 15.2 \\
$200$ & $84.6 \pm 2.5$ & 36.4 [3.7] & 2.17 & 1.00 & 11.5 \\
$400$ & $90.8 \pm 4.7$ & 36.8 [5.0] & 1.99 & 1.00 & 11.1 \\
\bottomrule
\end{tabular}
\end{table}

\section{Learning-Based Baselines}
\label{sec:appendix:baselines}

None of the three learning-based methods has an implementation for our environments, so we
reimplemented each on open tabletop, against REAP's symbolic action space (extended for BT-h,
see below) and feasibility-check interface. All three are evaluated on the same $300$ test problems with a
$300$\,s cap per problem. This appendix describes the main parts of each implementation and
where it differs from the original paper.

\paragraph{SAHS.} SAHS~\citep{kim2019learning,kim2022representation} learns two relational
GNNs, one for picks and one for puts, that score abstract actions over a state abstraction
with geometric predicates: whether an object can be grasped or placed, and which objects
occlude these motions. The search is a best-first search over abstract actions whose
priority combines a heuristic that counts the objects to move, following the occlusion
predicates recursively, with a softmax over the network scores. We follow the IJRR version,
with its full edge encoding and its large-margin ranking loss. In our adaptation, the objects
are the cylinders, and a put can target any empty cell, about $63$ per state. The geometric
predicates are computed with inverse kinematics and motion planning, and they account for
almost all of the motion-planner queries of SAHS. The networks are trained on one example per
step of the reference plans of REAP's $600$ training problems ($2{,}019$ examples from the
$573$ plans that could be replayed), with Adam (learning rate $10^{-3}$, full batch, $300$
epochs). Many put candidates have exactly the same input features as the chosen one; we
exclude these duplicates from the ranking losses of both networks, without which the put
network does not learn. We report one unseeded run with the final weights. Other differences from the paper: the arm has a fixed base and the predicates are
computed from the pick and place motions of the arm rather than from the swept volumes of a
mobile base, the heuristic drops the terms for goal regions since the goal (holding the
target) has none, and each pick attempt uses a single grasp. The feasibility check retries IK
from up to three seeds, uses more IK and planner iterations than REAP's, runs $16$ processes
per problem, and retries a failed action up to three times.

\paragraph{PLOI.} PLOI~\citep{silver2021planning} learns a GNN that scores the importance of
each object from the initial state and the goal. At test time, it plans on the subproblem
with the objects scoring at least $\gamma^N$, for $N = 1, 2, \ldots$, until the plan found
also solves the full problem. In our adaptation, the graph has a node for each cell and each
cylinder, with node features for the object type and for emptiness and holding in the initial
state and in the goal, and edges for occupancy and adjacency. Only cylinders are scored and
pruned, and goal objects are always kept. The planner is REAP's breadth-first search or
greedy best-first search with $h_{\text{FF}}$, with the same feasibility check, in place of
the paper's planners. As in the authors' code, and unlike the greedy backward removal
described in the paper, a cylinder is labeled important if a plan for the full problem,
found by breadth-first search with an $h_{\text{FF}}$ fallback, touches it; training problems
without such a plan are dropped, which leaves $436$ of the $600$. The network follows the paper
and the authors' code (encode-process-decode with $3$ message-passing rounds, hidden width
$16$, weighted cross-entropy) with $\gamma = 0.9$; it is trained from a single seed, with
early stopping on a validation set instead of a fixed number of epochs. Each planner call has
its own $300$\,s budget and the overall limit is checked only between calls, so a few problems
are solved after $300$\,s; Table~\ref{tab:opentabletop} counts them as unsolved. A variant with positional node features
does not perform better.

\paragraph{BT-h.} BT-h~\citep{cieslar2025learning} refines a given plan skeleton by
backtracking over the parameters of its actions. A transformer classifier predicts whether a
partially refined skeleton can be completed; candidates with low predicted feasibility are
discarded, and the confidence of the classifier decides when to jump back. Our open-tabletop
domain fixes the target cell in every action, which leaves nothing to refine, so we extend it
with a \texttt{relocate} action whose target cell is chosen during refinement, uniformly among
the free cells. At test time, BT-h is given the order in which the reference plan of the
problem moves the cylinders, and chooses the target cells. These reference plans were found
by our problem generator with a search that checks every action for feasibility; the cost of
that search is not counted in the results of BT-h. Since our pipeline does not convert REAP's
training problems to the extended domain, the classifier is trained with the paper's
data-collection procedure on synthetic problems: the $575$ of $600$ problems that the
demonstrations solve, with $2$ to $7$ thin cylinders, $1$ to $6$ of them blocking the target
from directly in front, in $48$ distinct layouts. Data collection uses $30$ search trees per
level, each keeping all its nodes as examples, instead of $4{,}000$ that keep only the first.
The classifier is smaller and trained for less than in the paper (width $32$, $4$ heads and
$2$ blocks instead of $128$, $8$ and $4$; batches of $32$ for at most $400$ steps per round
instead of $4{,}000$ for $5{,}000$), its veto threshold was set by hand, and we train a
single seed. The feasibility check runs
in a single process with a smaller planner step than REAP's and re-checks an infeasible
verdict up to two times. Given the same orderings, plain backtracking without the classifier
solves $85.0\%$ of the problems, against $81.7\%$ for BT-h.

\section{PDDLStream Baseline}
\label{sec:appendix:pddlstream}

PDDLStream~\citep{garrett2020pddlstream} is a task-and-motion planner that interleaves
symbolic search with calls to \emph{streams}, procedures that sample or test the continuous
parameters of actions. We run its adaptive algorithm on the same $300$ open-tabletop problems
as REAP, with the same scenes, the same feasibility-check interface and the same $300$\,s cap, which
does not include the construction of the scene (about $10$\,s). In our encoding, the streams
only check whether a pick or a put is feasible, so no continuous parameter is sampled, and a
motion-planner query is one such check.

We evaluate the four configurations of Table~\ref{tab:opentabletop}. In the two
\emph{test-stream} configurations, all grid cells are given in the initial state and a
fluent stream, an extension available in the PDDLStream code, tests whether a cylinder can
be put in a cell; Fast Downward runs in its \texttt{ff-astar2} or \texttt{ff-eager} mode. In
the \emph{cell-generator} configuration, cells are instead proposed lazily by a generator,
which keeps the optimistic action space of the planner small, and the fluent test still
checks every placement. The \emph{informed cells + occlusion axiom} configuration orders the
generated cells by the initial layout (free cells first, then cells with fewer occupied
neighbors and farther from the target) and adds a precondition to \texttt{pick}: no other
object may occlude the object to be picked, as given by a derived \emph{occluder} predicate.
It is the strongest of our formulations on relocation problems. The last two configurations use \texttt{ff-eager}.
All four use unit action costs, return the first plan found, use a search-to-sample ratio of
$1$, and give Fast Downward $30$\,s per call.

\paragraph{Where PDDLStream Fails.} Table~\ref{tab:appendix:pddlstream:refl} groups the
problems by the length of their reference plan; a reference plan of length $2k+1$ relocates
$k$ objects. All four configurations solve $98$ to $99\%$ of the problems without
relocation. The three configurations without the occlusion axiom solve at most $22\%$ of the
problems with one relocation and none with more; the cell generator helps only slightly.
Their symbolic search has no information about which object blocks the target, since
feasibility enters only through the streams. The occlusion axiom provides this information:
with it, PDDLStream solves $91\%$ of the problems with one relocation and $34\%$ of those with
two or three, but none with four or more. Table~\ref{tab:appendix:pddlstream:perdataset}
shows the same per evaluation set. PDDLStream thus fails where hidden geometric preconditions
force long sequences of relocations, the setting REAP is designed for, and a hand-written
axiom for these preconditions recovers only the shorter sequences.

\paragraph{Motion-Planner Queries.} The query counts of PDDLStream in
Table~\ref{tab:opentabletop} are averages over the problems each configuration solves, which
for the three configurations without the occlusion axiom are almost all problems without
relocation, so they are not directly comparable with those of REAP.
Table~\ref{tab:appendix:pddlstream:common} compares both on the problems that both solve,
where a problem counts as solved by REAP only if all five seeds solve it. For the two
test-stream configurations, these are almost all one-action problems: $124$ of the $125$
problems solved by \texttt{ff-astar2} are also solved by \textsc{REAP-GBFS}, and their mean
reference plan length is $1.1$. On them, \textsc{REAP-greedy} needs fewer queries than
PDDLStream ($1.3$ against $1.7$ for \texttt{ff-astar2}), while \textsc{REAP-GBFS} needs more
($3.6$ against $1.6$), since its search also expands alternatives when a single action
suffices. For the other two configurations the order is reversed: \textsc{REAP-GBFS} needs
about three times fewer queries than the cell generator ($3.6$ against $9.8$, over $131$
problems) and about four times fewer than the configuration with the occlusion axiom ($6.0$
against $23.1$, over $190$ problems); this configuration also solves harder problems (mean
reference length $2.1$).

\begin{table}[h]
\centering
\caption{Solve rate (\%) of PDDLStream by reference plan length, with the number of
relocations it requires.}
\label{tab:appendix:pddlstream:refl}
\small
\begin{tabular}{lccccc}
\toprule
Reference plan length & Problems & test, ff-astar2 & test, ff-eager & cell gen. & occl.\ axiom \\
\midrule
$1$ (no relocation)          & 122 & 98 & 99 & 99 & 99 \\
$3$ (one relocation)         & 46  & 13 & 15 & 22 & 91 \\
$5$--$7$ (two or three)      & 88  & 0  & 0  & 0  & 34 \\
$9$--$15$ (four or more)     & 44  & 0  & 0  & 0  & 0  \\
\bottomrule
\end{tabular}
\end{table}

\begin{table}[h]
\centering
\caption{Solve rate (\%) of PDDLStream per evaluation set ($20$ problems each). The
\texttt{ff-eager} test-stream configuration is within two problems of \texttt{ff-astar2} in
every set.}
\label{tab:appendix:pddlstream:perdataset}
\small
\begin{tabular}{lccc}
\toprule
Evaluation set & test, ff-astar2 & cell gen. & occl.\ axiom \\
\midrule
\multicolumn{4}{l}{\emph{Object count (random placement)}} \\
obj10 & 45 & 45 & 90 \\
obj12 & 75 & 75 & 95 \\
obj14 & 65 & 60 & 90 \\
obj16 & 80 & 80 & 95 \\
obj18 & 70 & 75 & 90 \\
obj20 & 85 & 100 & 100 \\
obj22 & 95 & 95 & 95 \\
obj24 & 90 & 95 & 95 \\
\midrule
\multicolumn{4}{l}{\emph{Plan length (random placement)}} \\
obj10\_plan3         & 5  & 10 & 55 \\
obj12\_plan3         & 15 & 20 & 55 \\
obj14\_plan5         & 0  & 0 & 30 \\
obj16\_plan5         & 0  & 0 & 40 \\
\midrule
\multicolumn{4}{l}{\emph{Arrangement ($K$ direct blockers)}} \\
obj10\_plan5\_biased2 & 0  & 0 & 35 \\
obj12\_plan7\_biased3 & 0  & 0 & 0 \\
obj14\_plan9\_biased3 & 0  & 0 & 0 \\
\bottomrule
\end{tabular}
\end{table}

\begin{table}[h]
\centering
\caption{Motion-planner queries per plan on the problems solved by both PDDLStream and REAP.
\emph{Ref.\ $|\pi|$}: mean reference plan length of the problems the configuration solves;
$|\text{both}|$: number of problems solved by both.}
\label{tab:appendix:pddlstream:common}
\small
\resizebox{\linewidth}{!}{%
\begin{tabular}{lccccccc}
\toprule
 & & \multicolumn{3}{c}{vs.\ \textsc{REAP-GBFS}} & \multicolumn{3}{c}{vs.\ \textsc{REAP-greedy}} \\
\cmidrule(lr){3-5}\cmidrule(lr){6-8}
PDDLStream configuration & Ref.\ $|\pi|$ & $|\text{both}|$ & PDDLStream & REAP & $|\text{both}|$ & PDDLStream & REAP \\
\midrule
test-stream, ff-astar2 & $1.1$ & $124$ & $1.6$ & $3.6$ & $115$ & $1.7$ & $1.3$ \\
test-stream, ff-eager & $1.1$ & $128$ & $2.1$ & $3.6$ & $119$ & $2.2$ & $1.3$ \\
cell-generator, ff-eager & $1.2$ & $131$ & $9.8$ & $3.6$ & $121$ & $10.5$ & $1.4$ \\
informed cells + occlusion axiom & $2.1$ & $190$ & $23.1$ & $6.0$ & $176$ & $23.8$ & $2.5$ \\
\bottomrule
\end{tabular}}
\end{table}

\section{Execution Under Perturbed Observations}
\label{sec:appendix:noise}

REAP assumes that the symbolic state is given rather than perceived. To test how much this
assumption matters, we execute the plans of a trained policy on the open-tabletop problems
under three kinds of perturbation. The policy is a checkpoint of the combined configuration
from an earlier training run, not one of the five seeds of Table~\ref{tab:opentabletop}; only
the execution is perturbed.

\begin{itemize}
  \item \textbf{Placement drift.} Each \texttt{put} places the cylinder with a Gaussian
        position error with standard deviation $\sigma \in \{2, 5, 10\}$\,mm.
  \item \textbf{Occupancy errors.} When the state is observed, the occupancy of each cell is
        flipped with probability $p \in \{0.01, 0.02, 0.05\}$.
  \item \textbf{Grasp failures.} Each grasp fails with probability
        $p \in \{0.02, 0.05, 0.10\}$.
\end{itemize}

When the observed state differs from the expected one, the system replans once from the
observed state. Table~\ref{tab:appendix:noise} reports the fraction of episodes that finish
without replanning and, among those that replan, the fraction that then succeed.

\begin{table}[h]
\centering
\caption{Execution under perturbation, \textsc{REAP-GBFS} / \textsc{REAP-greedy}.
\emph{Completed}: episodes finished without replanning; \emph{recovered}: episodes that
succeed after replanning, among those that replan.}
\label{tab:appendix:noise}
\small
\begin{tabular}{llcc}
\toprule
Perturbation & Level & Completed (\%) & Recovered (\%) \\
\midrule
\multicolumn{4}{l}{\emph{Placement drift}} \\
\quad $\sigma$ & $2$\,mm  & $85.4$ / $81.6$ & $54.8$ / $60.4$ \\
\quad          & $5$\,mm  & $71.5$ / $73.4$ & $14.6$ / $18.8$ \\
\quad          & $10$\,mm & $52.4$ / $50.8$ & $6.6$ / $7.8$ \\
\midrule
\multicolumn{4}{l}{\emph{Occupancy errors}} \\
\quad $p$ & $0.01$ & $86.8$ / $83.0$ & $100.0$ / $91.1$ \\
\quad     & $0.02$ & $85.7$ / $87.2$ & $100.0$ / $97.1$ \\
\quad     & $0.05$ & $80.8$ / $79.2$ & $100.0$ / $94.5$ \\
\midrule
\multicolumn{4}{l}{\emph{Grasp failures}} \\
\quad $p$ & $0.02$ & $89.5$ / $84.0$ & $100.0$ / $95.2$ \\
\quad     & $0.05$ & $86.4$ / $84.0$ & $97.4$ / $95.2$ \\
\quad     & $0.10$ & $75.3$ / $77.9$ & $91.5$ / $84.5$ \\
\bottomrule
\end{tabular}
\end{table}

\paragraph{Replanning Recovers Transient Errors Better than Persistent Ones.}
Of the episodes that replan, a single replan recovers $91$ to $100\%$ under occupancy errors
and $85$ to $100\%$ under grasp failures, but under placement drift only $55$ to $60\%$ at $2$\,mm and $7$ to $19\%$ at $5$ and $10$\,mm.
The difference follows from what each perturbation does. A wrongly observed cell or a failed
grasp is \emph{transient}: the next observation or the next attempt sees the true state, and
a plan from that state is as good as the original one. Drift is \emph{persistent}: the
cylinder really is a few millimeters away from the center of its cell, the grid abstraction
cannot represent this, and the new plan often runs into the same kind of failure.

The limitation therefore lies in the abstraction and not in the search: \textsc{REAP-GBFS}
and \textsc{REAP-greedy} degrade in much the same way under all three perturbations, although their
solve rates without perturbation differ by about eight points ($96.3\%$ against $88.0\%$).
Overcoming it would require an abstraction that represents continuous positions, or a
feasibility check that uses the observed rather than the assumed scene.

\end{document}